\documentclass[10pt,twocolumn,letterpaper]{article}

\usepackage[pagenumbers]{cvpr}
\usepackage{amsmath,amssymb}
\usepackage{graphicx}
\usepackage{booktabs,array,multirow}
\usepackage[table]{xcolor}
\usepackage{xspace}
\usepackage{microtype}
\usepackage{tikz}
\usepackage{placeins}
\usepackage{float}
\usepackage{adjustbox}
\usetikzlibrary{arrows.meta,positioning,calc,fit,backgrounds}

\newcommand{\livecue}{LiveCue\xspace}
\newcommand{\silence}{\texttt{Silence}\xspace}
\newcommand{\response}{\texttt{Response}\xspace}

\newcommand{\metricup}{\ensuremath{\uparrow}}
\newcommand{\metricdown}{\ensuremath{\downarrow}}

\definecolor{livecueblue}{RGB}{40,82,122}
\definecolor{livecuegreen}{RGB}{45,114,93}
\definecolor{livecuegray}{RGB}{238,241,244}
\definecolor{livecueorange}{RGB}{166,93,45}

\definecolor{ocrow}{HTML}{EDF5EE}
\definecolor{mute}{HTML}{8C857B}
\definecolor{deltaup}{HTML}{3F7F57}
\definecolor{deltadown}{HTML}{B5694F}
\newcommand{\up}[1]{{\fontsize{6}{7}\selectfont\color{deltaup}$\uparrow$#1}}
\newcommand{\down}[1]{{\fontsize{6}{7}\selectfont\color{deltadown}$\downarrow$#1}}

\newcommand{\dbox}[1]{\makebox[2.05em][l]{\hspace{1.2pt}#1}}

\usepackage{etoc}
\definecolor{ocgreen}{HTML}{4E8A64}
\definecolor{ocgreendeep}{HTML}{3F7352}
\usepackage[pagebackref,breaklinks,colorlinks,allcolors=ocgreendeep]{hyperref}
\hypersetup{
  pdftitle={LiveCue: Your VLM Is Secretly a Proactive Streaming Assistant},
  pdfauthor={Yuandong Yang, Xiangyu Zeng, Yuhan Zhu, Zhiqiu Zhang, Guowei Tang, Haoran Chen, Qingyi Si, Jiaqi Wang, Limin Wang}
}

\def\confName{CVPR}
\def\confYear{2027}

\makeatletter
\def\cvprsection{\@startsection {section}{1}{\z@}
   {-10pt plus -2pt minus -2pt}{7pt} {\large\bf\color{black}}}
\def\cvprsubsection{\@startsection {subsection}{2}{\z@}
   {-8pt plus -2pt minus -2pt}{5pt} {\elvbf\color{black}}}
\def\cvprsubsubsection{\@startsection {subsubsection}{3}{\z@}
   {-6pt plus -2pt minus -2pt}{3pt} {\tenbf\color{black}}}
\def\abstract{%
   \centerline{\large\bf\color{black} Abstract}%
   \vspace*{12pt}\noindent%
   \it\ignorespaces%
}
\makeatother
\newcommand{\livecuemark}[1][1em]{\begin{tikzpicture}[baseline=-0.36*#1,x=#1/58,y=#1/58]
  \draw[ocgreen,line width=3*#1/58] (0,0) circle (27.5);
  \draw[ocgreen,line width=6*#1/58,line cap=round] (46.5:13) arc[start angle=46.5,end angle=313.5,radius=13];
  \fill[ocgreen] (12.5,0) circle (3.6);
\end{tikzpicture}}

\title{\livecuemark[1.05em]\,\textcolor{ocgreen}{LiveCue}: Your VLM Is Secretly a Proactive Streaming Assistant}
\author{%
  \normalsize
  Yuandong Yang$^{1*}$\quad
  Xiangyu Zeng$^{1*}$\quad
  Yuhan Zhu$^{1}$\quad
  Zhiqiu Zhang$^{2,4}$\quad
  Guowei Tang$^{1}$\\[2pt]
  \normalsize Haoran Chen$^{1,2}$\quad
  Qingyi Si$^{3}$\quad
  Jiaqi Wang$^{3}$\quad
  Limin Wang$^{1\dagger}$\\[4pt]
  \small $^{1}$Nanjing University\quad
  $^{2}$Shanghai AI Laboratory\quad
  $^{3}$JD.com\quad
  $^{4}$Shanghai Jiao Tong University\\[3pt]
  \small $^{*}$Equal contribution\quad $^{\dagger}$Corresponding author\\[3pt]
  \small Project page: \href{https://mcg-nju.github.io/LiveCue}{\nolinkurl{https://mcg-nju.github.io/LiveCue}}
}

\begin{document}
\maketitle
\etocdepthtag.toc{mtmain}

\begin{abstract}
Streaming video assistants must follow ongoing user tasks and decide whether, when, and what to report as events unfold. These requirements differ from the query-driven generation paradigm of conventional vision-language models (VLMs), and recent approaches increasingly develop this capability through specialized streaming training. We revisit this trend and ask how far general-purpose VLMs can go without any additional training. We first provide a comprehensive analysis of the paradigm difference and of how these models behave when moving from query-driven responses to continuous proactive interaction. Our analysis shows that existing VLMs can already produce strong task-relevant answers, while struggling to translate this ability into reliable continuous interaction: their reports are not consistently shaped by user intent, task progress, and prior deliveries. Guided by this analysis, we introduce \textbf{LiveCue}, a \textbf{training-free harness} built from a few simple designs that organize the existing capabilities of frozen VLMs. LiveCue selects \textbf{reusable interaction specifications} according to user intent and \textbf{commits each delivered response to task-appropriate state} while pacing subsequent inference, guiding when to wait, respond, or provide an update. Without any training, LiveCue \textbf{leads all compared streaming-trained models and agent frameworks on six proactive interaction benchmarks}, and it improves consistently across three VLMs as the backbone becomes more capable. These findings show that general-purpose VLMs already hold much of what proactive interaction requires and lack mainly its organization; once organized, they form a strong and often overlooked basis against which the gains of specialized streaming training can be measured.
\end{abstract}

\section{Introduction}
\label{sec:introduction}
\begin{figure}[t]
\centering
\includegraphics[width=0.88\linewidth]{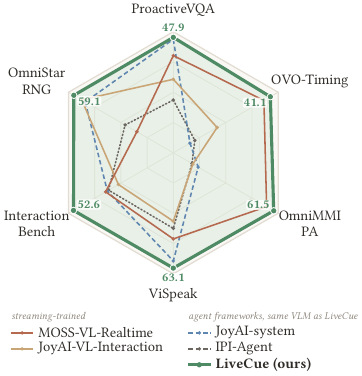}
\caption{\textbf{\livecue across six proactive benchmarks.} \livecue leads streaming-trained models and agent frameworks built on the same VLM on all six benchmarks, \textbf{without any additional training}. Each axis is rescaled to the best score shown; the numbers give \livecue's native scores.}
\label{fig:radar}
\end{figure}

\begin{figure*}[t]
\centering
\includegraphics[width=\textwidth]{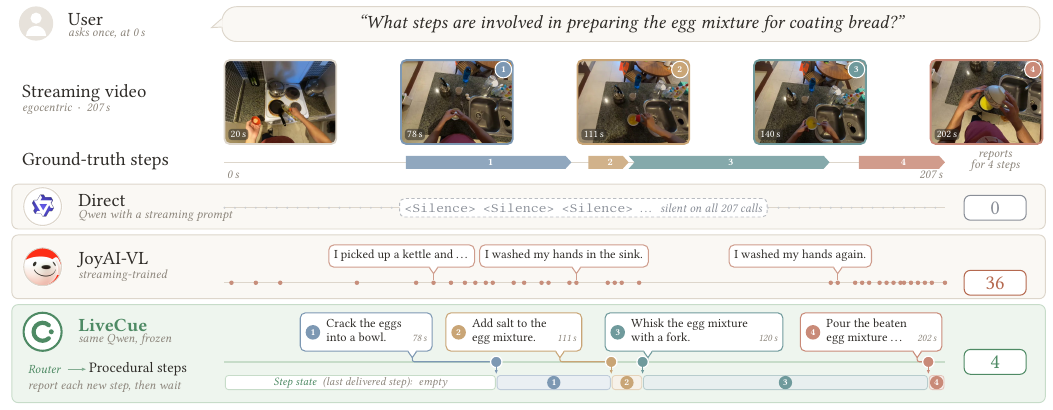}
\caption{\textbf{Organizing existing VLM capabilities for proactive interaction.} A user asks once for the steps of a procedure in an egocentric stream (ProactiveVideoQA). With a streaming prompt, Qwen3.8-27B stays silent on all 207 calls (Direct). The streaming-trained JoyAI-VL-Interaction narrates the scene with 36 reports. With the same frozen Qwen, \livecue routes the request to a procedural-steps protocol, keeps the last delivered step as task state, and reports each of the four steps once within its ground-truth window. All bubbles show real outputs; longer ones are shortened with an ellipsis.}
\label{fig:teaser}
\end{figure*}

Streaming video assistants are expected to support user tasks as events unfold. An assistant on AI glasses may guide an assembly procedure, watch for a requested event, or report an emerging anomaly. After specifying a task, the user expects the assistant to continue following the video without repeatedly asking for updates. Effective assistance requires the model to wait when evidence is insufficient, respond when the task warrants it, and provide further updates as relevant events develop. Such proactive video interaction therefore requires deciding whether, when, and what to report throughout an ongoing task.

These requirements extend beyond the typical query-driven use of vision-language models (VLMs)~\cite{qwen2vl,qwen25vl,qwen3vl,llavaonevision,internvl3}. In conventional video question answering, an explicit query specifies what the model should answer at the current turn. During proactive interaction, many subsequent updates contain new video frames without a new request, while the task remains in the preceding context. The model must interpret each update under this continuing instruction and determine whether it calls for a user-visible response. It must also consider task progress and information already delivered when deciding whether another report is needed. Correctly understanding the current scene is therefore only part of the problem: the system must turn that understanding into appropriate behavior over time.

Recent work has made substantial progress through specialized streaming training, which teaches models to coordinate waiting, response timing, and answer generation~\cite{videollm_online,aura,joyai,moss} and has become the mainstream route. Inference-time frameworks such as IPI-Agent~\cite{ipibench} have also explored proactive tasks with general-purpose models, yet how far a frozen VLM can go without training has received comparatively little attention. In the related setting of streaming video understanding, a deliberately simple recent-frame baseline has been shown to rival elaborate streaming designs~\cite{simplestream}. These developments raise a broader question: can frozen general-purpose VLMs achieve strong performance across diverse proactive tasks, and how should their existing capabilities be organized to reach it?

We investigate this question by analyzing the paradigm difference and directly deployed VLMs. On report windows where both respond, a directly prompted VLM produces answer content stronger than that of specially trained models, and when asked explicitly it recognizes nearly all of the alert targets that it fails to report during streaming. Its trajectories nevertheless miss later events, use report units that do not follow the task, and repeat reports. The capability for proactive assistance is thus already latent in the VLM; what it lacks is the organization to account for user intent, task progress, and prior reports in addition to visual evidence.

Guided by this analysis, we introduce \textbf{LiveCue}, a \textbf{training-free harness} that organizes the existing capabilities of frozen VLMs with a few deliberately simple designs (Figure~\ref{fig:teaser}). When a task is activated, LiveCue selects reusable interaction specifications according to the user's intent. These specifications define the relevant evidence, the unit of reporting, and the conditions for waiting or providing updates. As video arrives, the model checks each new observation against these specifications and makes an explicit decision to stay silent or respond, drawing on task state where needed, including progress and cumulative information for procedural and counting tasks. After a response is delivered, LiveCue commits it to the interaction history and task state and paces the next inference, so later decisions build on what has actually been reported. This process connects visual understanding with the continuing requirements of the interaction. It requires no additional model training and allows the same interaction framework to be reused across VLM backbones.

We evaluate LiveCue across six proactive interaction benchmarks covering diverse tasks and reporting requirements (Figure~\ref{fig:radar}). Without any training, LiveCue achieves the best score on every benchmark among the compared streaming-trained models and agent frameworks, and even on a 4B VLM its average score is comparable to that of the 11B MOSS-VL-Realtime~\cite{moss}. Across three Qwen VLMs spanning different generations and sizes, LiveCue achieves clear overall advantages over the evaluated JoyAI-system and IPI-Agent variants built on the same VLMs, and it substantially improves over the direct use of the same 27B model. More capable VLMs further improve overall proactive performance without additional training. Together, these results suggest that existing VLM capabilities can already support strong proactive interaction, and that training-free organization deserves to be considered alongside specialized streaming training.

Our contributions are threefold:
\begin{enumerate}
\item We analyze the paradigm difference between query-driven answering and continuous proactive interaction, revealing a gap between the answer competence of general-purpose VLMs and their proactive behavior.
\item We introduce LiveCue, a simple training-free harness that organizes existing VLM capabilities through \textbf{intent-aligned interaction protocols} with explicit response decisions, \textbf{task-appropriate state}, and \textbf{response commitment}.
\item Across six proactive benchmarks and three VLMs, we show that this training-free organization outperforms streaming-trained models and agent frameworks and improves consistently with the backbone.
\end{enumerate}

\section{Related Work}
\label{sec:related}

\subsection{Streaming and Proactive Video Interaction Benchmarks}

Early streaming video benchmarks, such as StreamingBench and OVO-Bench, already include initial tasks in which the model must decide when to respond as the video unfolds~\cite{streamingbench,ovobench}. Starting from ProactiveVideoQA~\cite{proactivevideoqa}, a series of benchmarks formally adopts native proactive evaluation and introduces metrics designed for proactive responses, including ViSpeak~\cite{vispeak}, OmniMMI~\cite{omnimmi}, and OmniStar~\cite{livestar}. InteractionBench further revisits the single-metric protocols of earlier benchmarks and jointly evaluates content, timing, and silence~\cite{interactionbench}. Across this progression, the tasks and scenarios become broader, and the evaluation settings and metrics become more complete, bringing assessment closer to real streaming interaction.

\subsection{Streaming-Trained Models for Proactive Interaction}

Video LLMs have advanced rapidly through large-scale video instruction tuning and long-context modeling~\cite{videollava,llavavideo,longva,videochatflash,internvideo25}, but they answer queries about a given clip. To equip VLMs with proactive responses, VideoLLM-online first introduces a streaming end-of-sequence decision learned on every frame~\cite{videollm_online}. MMDuet adopts a video--text duet interaction format~\cite{mmduet}, and Dispider and StreamBridge add separate decision modules to the video language model~\cite{dispider,streambridge}. Because silent moments vastly outnumber response moments in a stream, such supervision is highly imbalanced. Recent methods address this imbalance explicitly: Streamo applies focal weighting and state-frequency balancing~\cite{streamo}, AURA balances silent and non-silent supervision~\cite{aura}, JoyAI-VL-Interaction reweights response onsets and repeated silence~\cite{joyai}, and MMDuet2 optimizes the interaction policy with multi-turn reinforcement learning~\cite{mmduet2}. Together with streaming foundation models such as VideoChat3, Mage-VL, and MOSS-VL-Realtime~\cite{videochat3,mage,moss}, these methods clearly improve over earlier models. Complementary work improves memory and efficiency for long or unbounded streams~\cite{flashvstream,videochatonline,streamingvlm,timechatonline}. OneStreamer couples the two, learning query-independent caption memory and proactive response within a single streaming-trained model~\cite{onestreamer}. Streaming supervision has thus become the established route to proactive interaction. Its limitations are nevertheless notable. Performance on independent proactive benchmarks remains limited, particularly for tasks beyond the training data. Streaming specialization can also weaken general video understanding on benchmarks such as Video-MME, MVBench, LongVideoBench, and EgoSchema~\cite{videomme,mvbench,longvideobench,egoschema,aura,joyai,evostreaming,proactvl}, and the learned behavior must be retrained for each new backbone.

\subsection{Agent Frameworks for Proactive Video Interaction}

Organizing frozen models at inference time is well established for language agents, which interleave reasoning with actions and manage memory beyond the context window~\cite{react,memgpt}, and for offline video understanding, where agents iteratively search and store visual evidence~\cite{videoagent_wang,videoagent_fan}. For proactive interaction, only a few works follow this route. JoyAI-VL-Interaction provides an interaction framework designed together with its trained model~\cite{joyai}, and IPI-Agent combines a general MLLM with interaction memory, task management, and an embedding-based temporal gate, targeting tasks centered on its IPIBench~\cite{ipibench}. Applied to general-purpose VLMs across diverse proactive benchmarks, both yield limited performance (Sec.~\ref{sec:exp:main}). Whether general-purpose VLMs can support broad proactive interaction thus remains underexplored; we revisit this question by analyzing their behavior and organizing their existing capabilities with \livecue.

\section{Revisiting General VLMs for Proactive Interaction}
\label{sec:observations}

A proactive assistant must decide when a new event warrants a report,
what constitutes one report for the current request, and how later
observations relate to what has already been delivered. We formalize
this control problem and use a directly prompted general VLM to
separate answer competence from interaction organization.

\subsection{Proactive Interaction Adds a Control Problem}
\label{sec:obs:paradigm}

For a frozen VLM with parameters $\theta_0$, query-driven answering
and proactive interaction share the conditional-generation interface
\begin{equation}
 y_t \sim p_{\theta_0}(\cdot \mid H_{<t},v_t,q_t),
 \label{eq:shared}
\end{equation}
where $H_{<t}$ contains the task request and preceding interaction
history, $v_t$ is the visual content available at time $t$, and $q_t$
may be empty. In conventional VQA, the query specifies what should be
answered and $y_t=a_t$. In proactive interaction, most ticks bring only
new frames, so $q_t=\varnothing$ while the task in $H_{<t}$ remains
active, and each tick becomes a temporal control decision rather than
a new question. A state-first interface makes this decision explicit:
\begin{equation}
 y_t=\sigma_t\oplus a_t,\qquad
 \sigma_t\in\{\mathsf{Silence},\mathsf{Response}\},
 \label{eq:state-first}
\end{equation}
where $\oplus$ concatenates the control state and optional answer
content, and $a_t=\varnothing$ on the $\mathsf{Silence}$ branch.
Beyond what to answer, the model must decide whether to report and
whether the corresponding unit has already been delivered.

\subsection{Direct Inference Exposes a Capability--Organization Gap}
\label{sec:obs:diagnosis}

We run Qwen3.8-27B directly on the stream, prompting it at each step
to either stay silent or respond ($\silence/\response$), and refer to
this baseline as Direct. We evaluate it on all six benchmarks with
their native scorers and strict tag parsing.

\paragraph{Answers are strong when the model responds.}
On ground-truth windows answered by both Direct and a
streaming-trained model, Direct obtains a higher native content score
than each of five trained models on all four benchmarks with content
judgments (Table~\ref{tab:direct-content}). On OmniStar-RNG,
its advantage over JoyAI-VL-Interaction comes from semantic accuracy
(2.78 vs.~2.17) and completeness (2.92 vs.~2.05) rather than language
quality. Yet Direct answers only 25.2\% of OmniStar-RNG windows and
29.4\% of InteractionBench windows with content judgments.

\paragraph{Missed events are recognized on explicit query.}
To test whether low coverage reflects failed perception, we cut each
OmniMMI-PA video inside the target window, pair it with a cut before
the event, and ask the same VLM whether the requested event has
occurred. Qwen3.8-27B confirms 141 of the 143 target events and
answers 87.9\% of the paired prefixes correctly, above the 75.0\% of a
rule based on prefix duration alone. Of the 55 targets for which
Direct delivers no response inside the window during streaming, 54 are
confirmed. The missing reports thus
stem from turning recognition into a timely report rather than from
recognition itself.

\begin{figure*}[t]
\centering
\includegraphics[width=\textwidth]{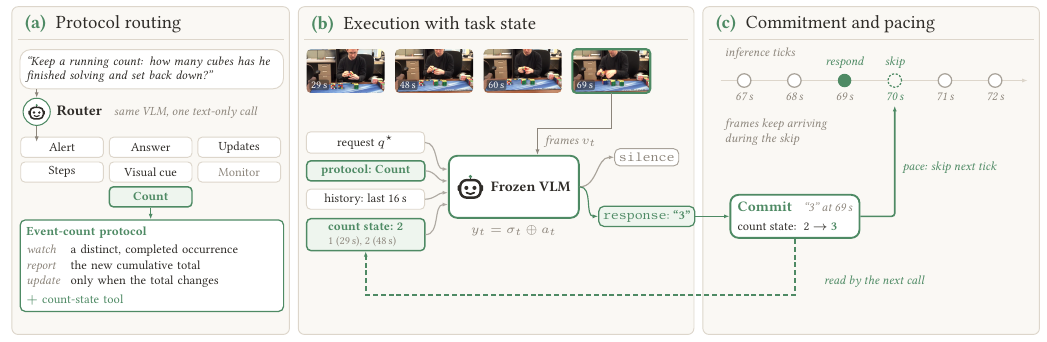}
\caption{\textbf{Overview of \livecue.} (a) When a task is activated, the same frozen VLM makes one text-only call that routes the request to an interaction protocol, which specifies the evidence to watch, the report unit, and when to update. (b) At each scheduled tick, the VLM decides between \silence and \response from the request, the protocol, the causal frames, the recent 16\,s history, and the task state; the count state retains earlier reports that are no longer in the history. (c) A delivered response is committed to the task state read by the next call and paces inference by skipping the next tick while frames keep arriving. The example is a counting request from InteractionBench with real \livecue outputs.}
\label{fig:harness}
\end{figure*}

\begin{table}[t]
\centering
\footnotesize
\setlength{\tabcolsep}{2.2pt}
\begin{tabular}{@{}lcccc@{}}
\toprule
Trained model & PVQA & ViSpeak & IB & RNG \\
 & \scriptsize(0--2) & \scriptsize(0--5) & \scriptsize(0--1) & \scriptsize(0--10) \\
\midrule
Mage-VL & \textbf{1.54}/0.81 & \textbf{4.00}/1.42 & \textbf{0.62}/0.49 & \textbf{4.49}/0.40 \\
JoyAI-VL-Int. & \textbf{1.41}/1.12 & \textbf{3.99}/3.72 & \textbf{0.61}/0.30 & \textbf{5.06}/4.54 \\
AURA & \textbf{1.41}/1.18 & \textbf{4.60}/3.38 & \textbf{0.51}/0.32 & -- \\
MOSS-VL-RT & \textbf{1.42}/1.19 & \textbf{4.15}/3.29 & \textbf{0.67}/0.32 & \textbf{4.79}/4.23 \\
VideoChat3 & \textbf{1.47}/0.94 & \textbf{3.99}/2.85 & \textbf{0.70}/0.48 & \textbf{4.81}/3.77 \\
\midrule
Direct coverage & 36.2\% & 67.8\% & 29.4\% & 25.2\% \\
\bottomrule
\end{tabular}
\caption{\textbf{Answer content on jointly answered windows.} Each cell gives Direct / the trained model on the ground-truth windows that both answer, scored by each benchmark's content judge (IB: non-multiple-choice items; RNG: SemCor). Unanswered windows are not scored. AURA is silent on RNG. The last row gives the fraction of windows Direct answers.}
\label{tab:direct-content}
\end{table}

\begin{table}[t]
\centering
\small
\setlength{\tabcolsep}{3.2pt}
\begin{tabular}{@{}lrrrr@{}}
\toprule
Benchmark & Coverage & Excess & Adjacent & Longest \\
 & \footnotesize(\%)\metricup & \footnotesize/100 GT\metricdown & \footnotesize(\%)\metricup & \footnotesize run\metricdown \\
\midrule
ProactiveVQA & 35.6 & 155.4 & 16.9 & 324 \\
OVO-Timing & 15.7 & 96.2 & 6.0 & 105 \\
OmniMMI-PA & 55.5 & 171.0 & -- & 139 \\
IB-A & 83.5 & 428.5 & -- & 11 \\
IB-B/C & 33.9 & 310.7 & 19.0 & 931 \\
OmniStar-RNG & 25.2 & 69.6 & 23.3 & 59 \\
\bottomrule
\end{tabular}
\caption{\textbf{Reporting behavior of Direct.} Coverage: ground-truth windows hit, with at most one hit per window. Excess: responses outside any window or beyond the first hit, per 100 windows. Adjacent: pairs of consecutive windows that are both hit (single-window benchmarks omitted). Longest run: most consecutive model calls that all respond.}
\label{tab:direct-behavior}
\end{table}

\paragraph{Interaction behavior is unreliable.}
Table~\ref{tab:direct-behavior} reveals three recurring patterns.
\emph{Report units do not follow user intent:} counting responses
outside any window or after its first hit as excess, Direct produces
155, 171, and 311 excess responses per 100 windows on
ProactiveVideoQA, OmniMMI-PA, and InteractionBench B/C, including 139
reports on one single-alert item and 97 on one four-step procedure.
\emph{Progress across events is lost:} only 6.0\% and 16.9\% of
adjacent windows are jointly covered on OVO-Timing and
ProactiveVideoQA, only 4 of 103 later procedure steps are reported,
and final counts are exact for only 13 of 160 InteractionBench items.
\emph{Delivered responses do not constrain later behavior:} misses and
excess reports co-occur, as Direct covers only 35.6\% of
ProactiveVideoQA windows at the excess rate above. Format errors do
not explain these failures; on trajectories with valid outputs only,
Direct still covers 26.7\% of OmniStar-RNG windows and answers the
first OmniMMI-PA trigger correctly for 48.7\% of items.

\paragraph{Implications.}
\label{sec:obs:implications}
Direct has the competence to answer but not the organization to
interact, and each failure pattern above points to a missing form of
support. An \emph{intent-aligned protocol} fixes what counts as a
report and when it is updated. \emph{Persistent task state} keeps
progress after the supporting frames leave the context.
\emph{Response commitment and pacing} record only delivered responses
in history and state, and keep the model from repeating a report it
has just made. \livecue{} provides these three forms of support around a
frozen VLM.

\section{LiveCue: Organizing Existing VLM Capabilities}
\label{sec:method}

\livecue organizes a frozen VLM around the three requirements of Sec.~\ref{sec:observations} (Figure~\ref{fig:harness}). It fixes an interaction protocol when a task begins (Sec.~\ref{sec:method:protocol}), runs continuous inference with task-appropriate state (Sec.~\ref{sec:method:state}), and commits each delivered response to guide later execution and pacing (Sec.~\ref{sec:method:feedback}), without updating any parameters. Each component relies on lightweight mechanisms, so the results reflect what a frozen VLM can already do once the interaction is organized.

\subsection{Intent-Aligned Interaction Protocols}
\label{sec:method:protocol}

A user request specifies both what to watch for and how to report it. ``Tell me when the button is pressed'' calls for a single notification once the condition holds, whereas ``update the count each time the button is pressed'' requires separating independent occurrences and reporting a running total. The two requests rely on the same visual evidence but demand different interactions. \livecue makes this requirement explicit as an \emph{interaction protocol}, which states the target evidence, the report unit, and the conditions for waiting, responding, and updating, and lets the model choose it when the task begins.

The protocol is paired with an explicit decision at every inference tick. Following the state-first interface of Eq.~\ref{eq:state-first}, the model generates
\begin{equation}
  y_t=\sigma_t\oplus a_t \sim p_{\theta_0}\!\left(\cdot \mid q^{\star}, P_z, v_t, H_{<t}, s_{<t}\right),
  \label{eq:decision}
\end{equation}
where $\sigma_t\in\{\mathsf{Silence},\mathsf{Response}\}$, $P_z$ is the selected protocol, $v_t$ is the visual context retained up to time $t$, $H_{<t}$ is the recent interaction history, $s_{<t}$ is the task state (Sec.~\ref{sec:method:state}), and $a_t$ is empty on the \silence branch. The decision is emitted through a structured output schema with constrained decoding, and a tolerant parser restores only unambiguous formatting. Each tick thus becomes a check of the newly arrived evidence against the protocol's criteria rather than a free continuation of the conversation: even when no new question arrives, the request and protocol define what to look for and when a report is due.

We distill a small library of reusable protocols from the task definitions of existing proactive and streaming benchmarks by asking what evidence to watch for, what constitutes one report, and when to wait or report again. The core protocols differ in their report unit: one-time alerts, evidence-conditioned answers, change-driven updates, procedural steps, event counts, and open-ended monitoring, complemented by protocols for responding to visual cues within a conversation. InteractionBench, the most comprehensive suite we evaluate, is held out from protocol design. The specific objects and events always come from the user's request.

When a task is activated, the same VLM that later performs the interaction makes one text-only call to select a protocol:
\begin{equation}
  z = C_{\theta_0}\!\left(q^{\star}, D\right),
  \label{eq:configure}
\end{equation}
where $q^{\star}$ is the standing request, $D$ is the visible dialogue context, and $z$ indexes the protocol library. This call decides only how to interact; whether a requested event has occurred is left to visual inference. Scope and reporting frequency stated explicitly by the user take precedence over the protocol's defaults. The selected protocol $P_z$ is reused for the duration of the task and selected again when the user changes the task. The protocol library is detailed in Appendix~\ref{sec:app:protocols}.

\begin{table*}[t]
\centering
\small
\setlength{\tabcolsep}{5.2pt}
\renewcommand{\arraystretch}{1.08}
\begin{tabular}{lrrrrrrr}
\toprule
\multirow{2}{*}{Model} & \multicolumn{1}{c}{Proactive} & \multicolumn{1}{c}{OVO-} & \multicolumn{1}{c}{OmniMMI} & \multicolumn{1}{c}{ViSpeak} & \multicolumn{1}{c}{Interaction} & \multicolumn{1}{c}{OmniStar} & \multicolumn{1}{c}{\multirow{2}{*}{Overall}} \\
 & \multicolumn{1}{c}{VideoQA} & \multicolumn{1}{c}{Timing} & \multicolumn{1}{c}{PA} & \multicolumn{1}{c}{} & \multicolumn{1}{c}{Bench} & \multicolumn{1}{c}{RNG} & \\
\midrule
\multicolumn{8}{l}{\textit{\color{mute}Streaming-trained models}} \\
Mage-VL~\cite{mage} & 25.7\dbox{} & 21.7\dbox{} & 5.5\dbox{} & 12.1\dbox{} & 31.0\dbox{} & 15.3\dbox{} & 18.5\dbox{} \\
JoyAI-VL-Interaction~\cite{joyai} & 32.3\dbox{} & 21.3\dbox{} & 16.0\dbox{} & 39.5\dbox{} & 31.2\dbox{} & 54.8\dbox{} & 32.5\dbox{} \\
AURA~\cite{aura} & 39.0\dbox{} & 11.7\dbox{} & 44.0\dbox{} & 15.6\dbox{} & 29.1\dbox{} & 0.0\dbox{} & 23.2\dbox{} \\
MOSS-VL-Realtime~\cite{moss} & 41.1\dbox{} & 38.7\dbox{} & 57.5\dbox{} & 48.4\dbox{} & 37.3\dbox{} & 24.8\dbox{} & 41.3\dbox{} \\
VideoChat3~\cite{videochat3} & 30.6\dbox{} & 33.6\dbox{} & 23.0\dbox{} & 18.3\dbox{} & 31.6\dbox{} & 34.5\dbox{} & 28.6\dbox{} \\
\midrule
\textbf{Qwen3.5-4B} & 37.1\dbox{} & 13.0\dbox{} & 17.0\dbox{} & 28.6\dbox{} & 19.4\dbox{} & 29.3\dbox{} & 24.1\dbox{} \\
\hspace{6pt}+ JoyAI system~\cite{joyai} & 29.2\dbox{\down{7.9}} & 19.6\dbox{\up{6.6}} & 18.5\dbox{\up{1.5}} & 44.2\dbox{\up{15.6}} & 26.2\dbox{\up{6.8}} & 21.1\dbox{\down{8.2}} & 26.4\dbox{\up{2.3}} \\
\hspace{6pt}+ IPI-Agent~\cite{ipibench} & 18.8\dbox{\down{18.3}} & 9.0\dbox{\down{4.0}} & 2.0\dbox{\down{15.0}} & 4.2\dbox{\down{24.4}} & 25.1\dbox{\up{5.7}} & 28.0\dbox{\down{1.3}} & 14.5\dbox{\down{9.6}} \\
\rowcolor{ocrow}\hspace{6pt}+ \livecue & \textbf{42.4}\dbox{\up{5.3}} & \textbf{25.2}\dbox{\up{12.2}} & \textbf{43.5}\dbox{\up{26.5}} & \textbf{51.6}\dbox{\up{23.0}} & \textbf{39.8}\dbox{\up{20.4}} & \textbf{56.0}\dbox{\up{26.7}} & \textbf{43.1}\dbox{\up{19.0}} \\
\midrule
\textbf{Qwen3.6-35B-A3B} & 36.3\dbox{} & 12.6\dbox{} & 18.0\dbox{} & 9.9\dbox{} & 25.0\dbox{} & 35.4\dbox{} & 22.8\dbox{} \\
\hspace{6pt}+ JoyAI system~\cite{joyai} & 45.4\dbox{\up{9.1}} & 16.3\dbox{\up{3.7}} & 1.5\dbox{\down{16.5}} & 30.1\dbox{\up{20.2}} & 21.5\dbox{\down{3.5}} & 50.6\dbox{\up{15.2}} & 27.6\dbox{\up{4.8}} \\
\hspace{6pt}+ IPI-Agent~\cite{ipibench} & 24.3\dbox{\down{12.0}} & 11.7\dbox{\down{0.9}} & 5.0\dbox{\down{13.0}} & 45.3\dbox{\up{35.4}} & 28.6\dbox{\up{3.6}} & 37.0\dbox{\up{1.6}} & 25.3\dbox{\up{2.5}} \\
\rowcolor{ocrow}\hspace{6pt}+ \livecue & \textbf{47.3}\dbox{\up{11.0}} & \textbf{32.3}\dbox{\up{19.7}} & \textbf{40.5}\dbox{\up{22.5}} & \textbf{58.9}\dbox{\up{49.0}} & \textbf{44.8}\dbox{\up{19.8}} & \textbf{53.1}\dbox{\up{17.7}} & \textbf{46.1}\dbox{\up{23.3}} \\
\midrule
\textbf{Qwen3.8-27B} & 41.9\dbox{} & 19.4\dbox{} & 49.5\dbox{} & 59.1\dbox{} & 40.9\dbox{} & 21.8\dbox{} & 38.8\dbox{} \\
\hspace{6pt}+ JoyAI system~\cite{joyai} & 47.2\dbox{\up{5.3}} & 11.2\dbox{\down{8.2}} & 19.0\dbox{\down{30.5}} & 59.5\dbox{\up{0.4}} & 36.5\dbox{\down{4.4}} & 52.9\dbox{\up{31.1}} & 37.7\dbox{\down{1.1}} \\
\hspace{6pt}+ IPI-Agent~\cite{ipibench} & 24.6\dbox{\down{17.3}} & 13.0\dbox{\down{6.4}} & 15.5\dbox{\down{34.0}} & 43.1\dbox{\down{16.0}} & 35.9\dbox{\down{5.0}} & 31.2\dbox{\up{9.4}} & 27.2\dbox{\down{11.6}} \\
\rowcolor{ocrow}\hspace{6pt}+ \livecue & \textbf{47.9}\dbox{\up{6.0}} & \textbf{41.1}\dbox{\up{21.7}} & \textbf{61.5}\dbox{\up{12.0}} & \textbf{63.1}\dbox{\up{4.0}} & \textbf{52.6}\dbox{\up{11.7}} & \textbf{59.1}\dbox{\up{37.3}} & \textbf{54.2}\dbox{\up{15.4}} \\
\bottomrule
\end{tabular}
\caption{\textbf{Proactive response performance across six benchmarks.} All evaluations use native streaming settings. For each general-purpose VLM, the first row is the VLM used directly with a streaming prompt, and arrows give each agentic system's change from it. Bold marks the best score per VLM. Overall averages the six benchmark scores; the RNG score is a balanced combination of its five native metrics.}
\label{tab:main-results}
\end{table*}

\subsection{Continuous Execution with Task State}
\label{sec:method:state}

Once the protocol is fixed, \livecue runs Eq.~\ref{eq:decision} at each scheduled tick on the standing request, the selected protocol, the causal visual context, and the interaction history. Fixing the protocol in advance also lets \livecue provide state targeted to the task. We equip procedural reporting and event counting with two compact state tools, whose records are supplied automatically as a short context block before each subsequent inference.

\paragraph{Step state.}
A procedural step often spans several observation ticks, and its result can remain visible after it is completed. The step tool keeps the most recently delivered step report and its video time. With this reference, the model can judge whether the current view still belongs to the reported step or shows a new step, a separate recurrence, or information that warrants a correction.

\paragraph{Count state.}
A cumulative count depends not only on the current event but on the total reported so far. The count tool keeps the latest parseable cumulative count together with the three most recent count reports and their video times, so the running total remains available after the corresponding frames and replies leave the recent context window. From new visual evidence, the model decides whether a distinct occurrence has happened and whether to keep, increment, or correct the total.

Both tools record what the model has actually delivered rather than ground truth. Event recognition, step judgment, and count revision remain the model's decisions, while the tools only store and supply the record.

\subsection{Response Commitment and Pacing}
\label{sec:method:feedback}

Let $r_t=1$ indicate that a response was actually delivered at tick $t$. The delivered text and its time are then committed as the basis for later execution: the recent history keeps the actual reply, and the applicable step or count state is updated,
\begin{equation}
  s_{t}=
  \begin{cases}
    U_z(s_{<t}, a_t, t), & r_t=1,\\
    s_{<t}, & r_t=0.
  \end{cases}
  \label{eq:state}
\end{equation}
Subsequent inference can therefore combine new observations with what has already been reported to decide whether another update is needed.

Commitment also paces inference. \livecue applies a brief cooldown after each delivered response. The scheduler skips the next inference tick, $g_{t+1}=1-r_t$, while frames continue to arrive, and then resumes model decisions. This limits back-to-back responses and lets the next decision see the newly arrived evidence. A skipped tick does not insert a synthetic message into the history.

The protocol library and pipeline are shared across VLMs; a thin adapter handles each model's visual input, history format, and decoding settings.

\section{Experiments}
\label{sec:results}

We compare \livecue with streaming-trained models and agent frameworks (Sec.~\ref{sec:exp:main}), ablate its three components (Sec.~\ref{sec:exp:ablation}), and follow its behavior across VLM backbones (Sec.~\ref{sec:exp:reuse}).

\subsection{Experimental Setup}
\label{sec:exp:setup}

\paragraph{Benchmarks and metrics.}
We evaluate on ProactiveVideoQA~\cite{proactivevideoqa}, OVO-Timing~\cite{ovobench}, OmniMMI-PA~\cite{omnimmi}, ViSpeak~\cite{vispeak}, InteractionBench~\cite{interactionbench}, and OmniStar-RNG~\cite{livestar}, covering 3,859 items. All evaluations use native streaming settings: frames arrive causally, and every response is scored at its actual video time with the benchmark's own metric and evaluator. We re-evaluate every system under these settings, with content judged by Qwen3.5-35B-A3B~\cite{qwen35} using benchmark-specific rubrics. We report one score per benchmark and their average as Overall. For OmniStar-RNG, whose five native metrics pull in opposite directions, we combine them into one balanced score defined in Appendix~\ref{sec:app:benchmarks}.
Per-task results for every benchmark are provided in Appendix~\ref{sec:app:perbench}.

\paragraph{Models and systems.}
\livecue is instantiated on Qwen3.5-4B~\cite{qwen35}, Qwen3.6-35B-A3B~\cite{qwen36}, and Qwen3.8-27B~\cite{qwen38} with the same protocol library and tool configuration. We compare against five streaming-trained models, two agent frameworks built on the same VLMs, JoyAI-system~\cite{joyai} and IPI-Agent~\cite{ipibench}, and the VLM used directly with a streaming prompt (Direct). Agent frameworks keep their own recipes, including auxiliary models; IPI-Agent, whose code is unreleased, is reimplemented from its paper.

\subsection{Main Results}
\label{sec:exp:main}

Table~\ref{tab:main-results} reports the six-benchmark results. Without additional training, \livecue on Qwen3.8-27B reaches 54.2 Overall and the best score on every benchmark among all compared streaming-trained models and agent frameworks. With Qwen3.5-4B, it reaches 43.1, comparable to the 41.3 of the 11B MOSS-VL-Realtime. On every VLM, \livecue obtains the highest score on all six benchmarks among the systems built on that VLM, and it improves the VLM used directly by 19.0, 23.3, and 15.4 Overall points on Qwen3.5-4B, Qwen3.6-35B-A3B, and Qwen3.8-27B, with gains on all 18 backbone--benchmark pairs. The two agent frameworks change Direct by only $-11.6$ to $+4.8$ points, and on Qwen3.8-27B both fall below it (37.7 and 27.2 against 38.8).

The streaming-trained models are not only weaker overall but highly uneven across benchmarks. MOSS-VL-Realtime reaches 57.5 on OmniMMI-PA but 24.8 on OmniStar-RNG, JoyAI-VL-Interaction shows the opposite pattern (16.0 and 54.8), and AURA stays silent throughout OmniStar-RNG. \livecue on Qwen3.8-27B scores at least 41.1 on every benchmark, a spread of 22.0 points against 32.7 for MOSS-VL-Realtime and 38.8 for JoyAI-VL-Interaction.

\paragraph{Where the gains come from.}
With the same weights, \livecue changes exactly the behaviors that Sec.~\ref{sec:obs:diagnosis} found unreliable in Direct. Coverage rises from 35.6\% to 63.2\% of ProactiveVideoQA windows and from 25.2\% to 95.1\% of OmniStar-RNG windows, while answer content stays above every streaming-trained model on jointly answered windows. Reports follow the requested unit: excess reports per 100 windows fall from 155.4 to 65.3 on ProactiveVideoQA and from 171.0 to 83.0 on OmniMMI-PA, and coverage of OVO-Timing steps after the first rises from 3.9\% to 64.1\%. On OmniMMI-PA, \livecue reports 34 of the 54 targets that Direct missed during streaming yet confirmed on explicit query.

\subsection{Ablation Studies}
\label{sec:exp:ablation}

\begin{table}[t]
\centering
\footnotesize
\setlength{\tabcolsep}{2.4pt}
\begin{adjustbox}{max width=\linewidth}
\begin{tabular}{@{}lccccccc@{}}
\toprule
System & PVQA & OVO & PA & ViSp. & IB & RNG & Overall \\
\midrule
Direct, free-form & \textbf{51.3} & 8.1 & 0.0 & 14.6 & 34.0 & 54.6 & 27.1 \\
Direct, \silence/\response & 41.9 & 19.4 & 49.5 & 59.1 & 40.9 & 21.8 & 38.8 \\
\midrule
+ Protocol, decision & 47.3 & 32.6 & 59.0 & 59.9 & 49.3 & 54.7 & 50.5 \\
+ Commit, pacing & 47.6 & 31.9 & 59.0 & 59.9 & 49.9 & \textbf{59.1} & 51.2 \\
\rowcolor{ocrow}+ Routing, state (\livecue) & 47.9 & \textbf{41.1} & \textbf{61.5} & \textbf{63.1} & \textbf{52.6} & \textbf{59.1} & \textbf{54.2} \\
\bottomrule
\end{tabular}
\end{adjustbox}
\caption{\textbf{From Direct to \livecue} on Qwen3.8-27B. Direct is prompted for streaming with free-form output or with the \silence/\response interface of Table~\ref{tab:main-results}. \livecue's components are then added cumulatively: a general interaction protocol with explicit structured decisions, response commitment with one-tick pacing, and intent-aligned routing with task state. PVQA: ProactiveVideoQA; PA: OmniMMI-PA; ViSp.: ViSpeak; IB: InteractionBench; RNG: OmniStar-RNG.}
\label{tab:abl-ladder}
\end{table}

\paragraph{From Direct to \livecue.}
Table~\ref{tab:abl-ladder} builds \livecue from the same frozen Qwen3.8-27B. With free-form output, the VLM speaks at nearly every step: it covers most ProactiveVideoQA and OmniStar-RNG windows but never places its first OmniMMI-PA report inside the target window. The \silence/\response interface makes it selective but leaves the failures of Sec.~\ref{sec:obs:diagnosis}. The largest step comes from the simplest change, stating the interaction as a protocol and deciding against it at every tick: 11.7 Overall points, with OVO-Timing rising from 19.4 to 32.6, OmniMMI-PA from 49.5 to 59.0, InteractionBench from 40.9 to 49.3, and OmniStar-RNG from 21.8 to 54.7. The visual competence is already present; what the VLM lacks is an explicit criterion for when its knowledge should become a report, which echoes the diagnosis of Sec.~\ref{sec:observations}. Commitment and pacing add 0.8 points, chiefly on OmniStar-RNG (54.7 to 59.1), by preventing the same observation from being reported on consecutive ticks. Routing to intent-aligned protocols with task state adds a further 3.0 points, concentrated where a single general protocol misreads the report unit: OVO-Timing gains 9.2, ViSpeak 3.2, InteractionBench 2.8, and OmniMMI-PA 2.5. Appendix~\ref{sec:app:ablation} compares routing with each protocol fixed for all tasks and tests its stability to rewording.

\begin{table}[t]
\centering
\footnotesize
\setlength{\tabcolsep}{2.0pt}
\begin{tabular}{@{}lcccccccc@{}}
\toprule
 & \multicolumn{4}{c}{With state tool} & \multicolumn{3}{c}{Without tool} & \\
\cmidrule(lr){2-5}\cmidrule(lr){6-8}
State given to & EGO & SSR & REC & IB-C & CRR & IB-B & PA & Overall \\
\midrule
None & 42.0 & 51.5 & 30.5 & 27.0 & \textbf{37.9} & \textbf{50.5} & \textbf{69.4} & 52.6 \\
All tasks & \textbf{44.3} & \textbf{63.7} & \textbf{41.5} & \textbf{34.9} & 29.2 & 44.5 & 62.6 & 53.0 \\
\rowcolor{ocrow}Routed (\livecue) & \textbf{44.3} & \textbf{63.7} & \textbf{41.5} & \textbf{34.9} & \textbf{37.9} & \textbf{50.5} & \textbf{69.4} & \textbf{54.2} \\
\bottomrule
\end{tabular}
\caption{\textbf{Ablation of persistent task state} on Qwen3.8-27B. Left: items routed to the step or count tool (EGO: ProactiveVideoQA-EGO pAUC; SSR: OVO-Timing recall; REC: exact final counts on OVO-Timing; IB-C: InteractionBench counting score). Right: items routed to state-free protocols (CRR: OVO-Timing F1; IB-B: InteractionBench delayed triggers; PA: OmniMMI-PA event F1).}
\label{tab:abl-state}
\end{table}

\paragraph{Persistent task state.}
We remove the step and count tools from the tasks routed to them, and conversely supply a record of the latest delivered response to every task routed to a state-free protocol (Table~\ref{tab:abl-state}). Removing the tools hurts exactly the tasks that track progress: procedural EGO pAUC drops from 44.3 to 42.0, SSR step recall from 63.7\% to 51.5\%, exact final counts on OVO-Timing from 34 to 25 of 82, and the InteractionBench counting score from 34.9 to 27.0. Supplying state everywhere leaves the routed tasks unchanged but lowers tasks with independent reports, with CRR falling from 37.9 to 29.2 and InteractionBench delayed triggers from 50.5 to 44.5. State helps where routing attaches it.

\subsection{\livecue Evolves with the Backbone}
\label{sec:exp:reuse}
\begin{figure}[t]
\centering
\includegraphics[width=\linewidth]{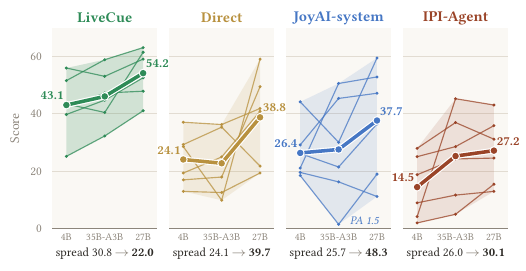}
\caption{\textbf{Per-benchmark evolution across VLM backbones.} Thin lines track each of the six benchmarks from Qwen3.5-4B to Qwen3.6-35B-A3B and Qwen3.8-27B, the shaded band spans their range, and the thick line gives Overall. The \livecue bundle stays high and rises together, narrowing as the VLM improves, whereas the same VLMs used directly (Direct) and the agent frameworks cross and drop, and their spread widens.}
\label{fig:evolve}
\end{figure}

The same protocol library and pipeline transfer unchanged across the three VLMs. Figure~\ref{fig:evolve} follows every benchmark across backbones for each system. The six benchmarks of \livecue rise together rather than only on average: from Qwen3.5-4B to Qwen3.8-27B every benchmark improves, and the only two declines between consecutive backbones are at most 3.0 points. The weakest benchmark rises from 25.2 to 32.3 to 41.1, so the spread narrows from 30.8 to 22.0 points. This consistency does not come from the VLM alone: used directly, the same VLMs widen their spread from 24.1 to 39.7 points. The agent frameworks also become less even. JoyAI-system falls to 1.5 on OmniMMI-PA with Qwen3.6-35B-A3B and its spread grows from 25.7 to 48.3, while IPI-Agent improves from a low base, with its weakest benchmark never above 13.0. Each advance in general-purpose VLMs thus yields consistent gains without streaming training.

\subsection{Efficiency}
\label{sec:exp:efficiency}

\livecue adds little inference cost to the VLM it organizes. Measured on 18 videos from the six benchmarks with NVIDIA H200 GPUs, \livecue on Qwen3.8-27B runs on the same single GPU with the same memory as Direct, adds only about 20 context tokens per call, and needs 0.58\,s per decision on average, comfortably within the 1\,Hz decision interval and faster than Direct (0.64\,s). On Qwen3.5-4B it needs 20.3\,GB and 0.19\,s per decision, while the 11B MOSS-VL-Realtime occupies 148.5\,GB. JoyAI-system, by contrast, requires a second GPU for its memory model and twice the context. Under real-time playback, with frames released on the video clock, \livecue on Qwen3.5-4B completes every decision within one second, including two continuous 300-second streams, and on Qwen3.8-27B 98\% of them, with the queue always recovering.

\section{Discussion and Conclusion}
\label{sec:conclusion}

We revisited the training-centric route to proactive video interaction and found that a general-purpose VLM is already a capable proactive assistant in latent form: it answers well and recognizes the requested events. Our analysis showed that turning this capability into reliable proactive execution requires explicit interaction specifications, task state, and response commitment. \livecue provides these through a training-free harness, and experiments across six benchmark families and three VLMs establish strong performance, gains from organization, and reuse across models. These results suggest that \textbf{training-free organization of general-purpose VLMs is a strong baseline for proactive interaction}, and that the gains of specialized streaming training are best measured against it under the same protocol.

% References follow the main text directly; the appendix starts on a new page.
{
  \small
  \bibliographystyle{ieeenat_fullname}
  \bibliography{references}

\begin{thebibliography}{49}
\providecommand{\natexlab}[1]{#1}
\providecommand{\url}[1]{\texttt{#1}}
\expandafter\ifx\csname urlstyle\endcsname\relax
  \providecommand{\doi}[1]{doi: #1}\else
  \providecommand{\doi}{doi: \begingroup \urlstyle{rm}\Url}\fi

\bibitem[Bai et~al.(2025{\natexlab{a}})Bai, Cai, Chen, Chen, Chen, Cheng, Deng,
  Ding, Gao, Ge, et~al.]{qwen3vl}
Shuai Bai, Yuxuan Cai, Ruizhe Chen, Keqin Chen, Xionghui Chen, Zesen Cheng,
  Lianghao Deng, Wei Ding, Chang Gao, Chunjiang Ge, et~al.
\newblock {Qwen3-VL Technical Report}.
\newblock \emph{arXiv preprint arXiv:2511.21631}, 2025{\natexlab{a}}.

\bibitem[Bai et~al.(2025{\natexlab{b}})Bai, Chen, Liu, Wang, Ge, Song, Dang,
  Wang, Wang, Tang, et~al.]{qwen25vl}
Shuai Bai, Keqin Chen, Xuejing Liu, Jialin Wang, Wenbin Ge, Sibo Song, Kai
  Dang, Peng Wang, Shijie Wang, Jun Tang, et~al.
\newblock {Qwen2.5-VL Technical Report}.
\newblock \emph{arXiv preprint arXiv:2502.13923}, 2025{\natexlab{b}}.

\bibitem[Chen et~al.(2024)Chen, Lv, Wu, Lin, Song, Gao, Liu, Gao, Mao, and
  Shou]{videollm_online}
Joya Chen, Zhaoyang Lv, Shiwei Wu, Kevin~Qinghong Lin, Chenan Song, Difei Gao,
  Jia-Wei Liu, Ziteng Gao, Dongxing Mao, and Mike~Zheng Shou.
\newblock {VideoLLM-online: Online Video Large Language Model for Streaming
  Video}.
\newblock In \emph{Proceedings of the IEEE/CVF Conference on Computer Vision
  and Pattern Recognition}, 2024.

\bibitem[{DeepSeek-AI}(2026)]{deepseekv41}
{DeepSeek-AI}.
\newblock {DeepSeek-V4.1-Flash: Pushing the Limits of KV Cache Compression}.
\newblock \emph{arXiv preprint arXiv:2609.19969}, 2026.

\bibitem[Fan et~al.(2024)Fan, Ma, Wu, Du, Li, Gao, and Li]{videoagent_fan}
Yue Fan, Xiaojian Ma, Rujie Wu, Yuntao Du, Jiaqi Li, Zhi Gao, and Qing Li.
\newblock {VideoAgent: A Memory-augmented Multimodal Agent for Video
  Understanding}.
\newblock \emph{arXiv preprint arXiv:2403.11481}, 2024.

\bibitem[Fu et~al.(2025{\natexlab{a}})Fu, Dai, Luo, Li, Ren, Zhang, Wang, Zhou,
  Shen, Zhang, et~al.]{videomme}
Chaoyou Fu, Yuhan Dai, Yongdong Luo, Lei Li, Shuhuai Ren, Renrui Zhang, Zihan
  Wang, Chenyu Zhou, Yunhang Shen, Mengdan Zhang, et~al.
\newblock {Video-MME: The First-Ever Comprehensive Evaluation Benchmark of
  Multi-modal LLMs in Video Analysis}.
\newblock In \emph{Proceedings of the IEEE/CVF Conference on Computer Vision
  and Pattern Recognition}, 2025{\natexlab{a}}.

\bibitem[Fu et~al.(2025{\natexlab{b}})Fu, Yang, Li, Peng, Lin, Wei, Hu, Xie,
  and Zheng]{vispeak}
Shenghao Fu, Qize Yang, Yuan-Ming Li, Yi-Xing Peng, Kun-Yu Lin, Xihan Wei,
  Jian-Fang Hu, Xiaohua Xie, and Wei-Shi Zheng.
\newblock {ViSpeak: Visual Instruction Feedback in Streaming Videos}.
\newblock In \emph{Proceedings of the IEEE/CVF International Conference on
  Computer Vision}, pages 21778--21788, 2025{\natexlab{b}}.

\bibitem[Huang et~al.(2025)Huang, Li, Li, Wang, Zeng, Liang, Wu, Chen, Li, and
  Wang]{videochatonline}
Zhenpeng Huang, Xinhao Li, Jiaqi Li, Jing Wang, Xiangyu Zeng, Cheng Liang, Tao
  Wu, Xi Chen, Liang Li, and Limin Wang.
\newblock {Online Video Understanding: OVBench and VideoChat-Online}.
\newblock In \emph{Proceedings of the IEEE/CVF Conference on Computer Vision
  and Pattern Recognition}, 2025.

\bibitem[Li et~al.(2025)Li, Zhang, Guo, Zhang, Li, Zhang, Zhang, Zhang, Li,
  Liu, et~al.]{llavaonevision}
Bo Li, Yuanhan Zhang, Dong Guo, Renrui Zhang, Feng Li, Hao Zhang, Kaichen
  Zhang, Peiyuan Zhang, Yanwei Li, Ziwei Liu, et~al.
\newblock {LLaVA-OneVision: Easy Visual Task Transfer}.
\newblock \emph{Transactions on Machine Learning Research}, 2025.

\bibitem[Li et~al.(2026{\natexlab{a}})Li, Chen, Song, Lei, Zhang, Yan, Pan, and
  Liu]{ipibench}
Jinzhao Li, Yinuo Chen, Wenxuan Song, Yijia Lei, Yichi Zhang, Honglei Yan,
  Panwang Pan, and Miao Liu.
\newblock {IPIBench: Evaluating Interactive Proactive Intelligence of MLLMs
  under Continuous Streams}.
\newblock \emph{arXiv preprint arXiv:2605.27074}, 2026{\natexlab{a}}.

\bibitem[Li et~al.(2024)Li, Wang, He, Li, Wang, Liu, Wang, Xu, Chen, Luo,
  et~al.]{mvbench}
Kunchang Li, Yali Wang, Yinan He, Yizhuo Li, Yi Wang, Yi Liu, Zun Wang, Jilan
  Xu, Guo Chen, Ping Luo, et~al.
\newblock {MVBench: A Comprehensive Multi-modal Video Understanding Benchmark}.
\newblock In \emph{Proceedings of the IEEE/CVF Conference on Computer Vision
  and Pattern Recognition}, 2024.

\bibitem[Li et~al.(2026{\natexlab{b}})Li, Wang, Yu, Zeng, Zhu, Huang, Gao, Li,
  He, Wang, et~al.]{videochatflash}
Xinhao Li, Yi Wang, Jiashuo Yu, Xiangyu Zeng, Yuhan Zhu, Haian Huang, Jianfei
  Gao, Kunchang Li, Yinan He, Chenting Wang, et~al.
\newblock {VideoChat-Flash: Hierarchical Compression for Long-Context Video
  Modeling}.
\newblock In \emph{International Conference on Learning Representations},
  2026{\natexlab{b}}.

\bibitem[Li et~al.(2026{\natexlab{c}})Li, Zhu, Zeng, Dong, Wu, Zhang, Yang, Ma,
  Zhang, Shi, et~al.]{videochat3}
Xinhao Li, Yuhan Zhu, Xiangyu Zeng, Yuhao Dong, Haoning Wu, Zhiqiu Zhang,
  Yuandong Yang, Changlian Ma, Qingyu Zhang, Yansong Shi, et~al.
\newblock {VideoChat3: Fully Open Video MLLM for Efficient and Generalist Video
  Understanding}.
\newblock \emph{arXiv preprint arXiv:2607.14935}, 2026{\natexlab{c}}.

\bibitem[Lin et~al.(2024{\natexlab{a}})Lin, Ye, Zhu, Cui, Ning, Jin, and
  Yuan]{videollava}
Bin Lin, Yang Ye, Bin Zhu, Jiaxi Cui, Munan Ning, Peng Jin, and Li Yuan.
\newblock {Video-LLaVA: Learning United Visual Representation by Alignment
  Before Projection}.
\newblock In \emph{Proceedings of the Conference on Empirical Methods in
  Natural Language Processing}, 2024{\natexlab{a}}.

\bibitem[Lin et~al.(2024{\natexlab{b}})Lin, Fang, Chen, Wan, Luo, Li, Liu, and
  Sun]{streamingbench}
Junming Lin, Zheng Fang, Chi Chen, Zihao Wan, Fuwen Luo, Peng Li, Yang Liu, and
  Maosong Sun.
\newblock {StreamingBench: Assessing the Gap for MLLMs to Achieve Streaming
  Video Understanding}.
\newblock \emph{arXiv preprint arXiv:2411.03628}, 2024{\natexlab{b}}.

\bibitem[Lu et~al.(2026)Lu, Bo, Chen, Li, Guo, Guan, Liu, Xu, Sun, Sun, Liu,
  and Li]{aura}
Xudong Lu, Yang Bo, Jinpeng Chen, Shuhan Li, Xintong Guo, Huankang Guan, Fang
  Liu, Dunyuan Xu, Peiwen Sun, Heyang Sun, Rui Liu, and Hongsheng Li.
\newblock {AURA: Always-On Understanding and Real-Time Assistance via Video
  Streams}.
\newblock \emph{arXiv preprint arXiv:2604.04184}, 2026.

\bibitem[Mangalam et~al.(2023)Mangalam, Akshulakov, and Malik]{egoschema}
Karttikeya Mangalam, Raiymbek Akshulakov, and Jitendra Malik.
\newblock {EgoSchema: A Diagnostic Benchmark for Very Long-form Video Language
  Understanding}.
\newblock In \emph{Advances in Neural Information Processing Systems}, 2023.

\bibitem[Niu et~al.(2025)Niu, Li, Miao, Ge, Zhou, He, Dong, Duan, Ding, Qian,
  Zhang, Zang, Cao, He, and Wang]{ovobench}
Junbo Niu, Yifei Li, Ziyang Miao, Chunjiang Ge, Yuanhang Zhou, Qihao He, Xiaoyi
  Dong, Haodong Duan, Shuangrui Ding, Rui Qian, Pan Zhang, Yuhang Zang, Yuhang
  Cao, Conghui He, and Jiaqi Wang.
\newblock {OVO-Bench: How Far is Your Video-LLMs from Real-World Online Video
  Understanding?}
\newblock In \emph{Proceedings of the IEEE/CVF Conference on Computer Vision
  and Pattern Recognition}, pages 18902--18913, 2025.

\bibitem[Packer et~al.(2023)Packer, Wooders, Lin, Fang, Patil, Stoica, and
  Gonzalez]{memgpt}
Charles Packer, Sarah Wooders, Kevin Lin, Vivian Fang, Shishir~G. Patil, Ion
  Stoica, and Joseph~E. Gonzalez.
\newblock {MemGPT: Towards LLMs as Operating Systems}.
\newblock \emph{arXiv preprint arXiv:2310.08560}, 2023.

\bibitem[Qian et~al.(2025)Qian, Ding, Dong, Zhang, Zang, Cao, Lin, and
  Wang]{dispider}
Rui Qian, Shuangrui Ding, Xiaoyi Dong, Pan Zhang, Yuhang Zang, Yuhang Cao,
  Dahua Lin, and Jiaqi Wang.
\newblock {Dispider: Enabling Video LLMs with Active Real-Time Interaction via
  Disentangled Perception, Decision, and Reaction}.
\newblock In \emph{Proceedings of the IEEE/CVF Conference on Computer Vision
  and Pattern Recognition}, 2025.

\bibitem[{Qwen Team}(2026{\natexlab{a}})]{qwen35}
{Qwen Team}.
\newblock {Qwen3.5: Towards Native Multimodal Agents}.
\newblock \url{https://qwen.ai/blog?id=qwen3.5}, 2026{\natexlab{a}}.

\bibitem[{Qwen Team}(2026{\natexlab{b}})]{qwen36}
{Qwen Team}.
\newblock {Qwen3.6-35B-A3B: Agentic Coding Power, Now Open to All}.
\newblock \url{https://qwen.ai/blog?id=qwen3.6-35b-a3b}, 2026{\natexlab{b}}.

\bibitem[{Qwen Team}(2026{\natexlab{c}})]{qwen38}
{Qwen Team}.
\newblock {Qwen3.8-27B}.
\newblock \url{https://huggingface.co/Qwen/Qwen3.8-27B}, 2026{\natexlab{c}}.

\bibitem[Shen et~al.(2026)Shen, Tian, Yang, and Liu]{simplestream}
Yujiao Shen, Shulin Tian, Jingkang Yang, and Ziwei Liu.
\newblock {A Simple Baseline for Streaming Video Understanding}.
\newblock \emph{arXiv preprint arXiv:2604.02317}, 2026.

\bibitem[Song et~al.(2026)Song, Yu, Xu, Su, Xu, Chai, Tang, Deng, Xu, and
  Gu]{interactionbench}
Enxin Song, Suhao Yu, Yifei Xu, Barbara Su, Weili Xu, Wenhao Chai, Yao Tang,
  Jie Deng, Haiyang Xu, and Jiatao Gu.
\newblock {InteractionBench: A Real-Time Interaction Benchmark for Streaming
  Video Systems}.
\newblock \emph{arXiv preprint arXiv:2610.05775}, 2026.

\bibitem[Wang et~al.(2025{\natexlab{a}})Wang, Feng, Lai, Xu, Li, Ge, Dehghan,
  Cao, and Huang]{streambridge}
Haibo Wang, Bo Feng, Zhengfeng Lai, Mingze Xu, Shiyu Li, Weifeng Ge, Afshin
  Dehghan, Meng Cao, and Ping Huang.
\newblock {StreamBridge: Turning Your Offline Video Large Language Model into a
  Proactive Streaming Assistant}.
\newblock In \emph{Advances in Neural Information Processing Systems},
  2025{\natexlab{a}}.

\bibitem[Wang et~al.(2024{\natexlab{a}})Wang, Bai, Tan, Wang, Fan, Bai, Chen,
  Liu, Wang, Ge, et~al.]{qwen2vl}
Peng Wang, Shuai Bai, Sinan Tan, Shijie Wang, Zhihao Fan, Jinze Bai, Keqin
  Chen, Xuejing Liu, Jialin Wang, Wenbin Ge, et~al.
\newblock {Qwen2-VL: Enhancing Vision-Language Model's Perception of the World
  at Any Resolution}.
\newblock \emph{arXiv preprint arXiv:2409.12191}, 2024{\natexlab{a}}.

\bibitem[Wang et~al.(2026{\natexlab{a}})Wang, Tan, Zhou, Zhou, Chen, He, Zeng,
  Cheng, Wang, Qian, et~al.]{moss}
Pengyu Wang, Chenkun Tan, Shaojun Zhou, Qirui Zhou, Yanxin Chen, Xingyang He,
  Huazheng Zeng, Jijun Cheng, Chenghao Wang, Xiaomeng Qian, et~al.
\newblock {MOSS-VL Technical Report}.
\newblock \emph{arXiv preprint arXiv:2608.15045}, 2026{\natexlab{a}}.

\bibitem[Wang et~al.(2024{\natexlab{b}})Wang, Zhang, Zohar, and
  Yeung-Levy]{videoagent_wang}
Xiaohan Wang, Yuhui Zhang, Orr Zohar, and Serena Yeung-Levy.
\newblock {VideoAgent: Long-form Video Understanding with Large Language Model
  as Agent}.
\newblock In \emph{Proceedings of the European Conference on Computer Vision},
  2024{\natexlab{b}}.

\bibitem[Wang et~al.(2025{\natexlab{b}})Wang, Li, Yan, He, Yu, Zeng, Wang, Ma,
  Huang, Gao, et~al.]{internvideo25}
Yi Wang, Xinhao Li, Ziang Yan, Yinan He, Jiashuo Yu, Xiangyu Zeng, Chenting
  Wang, Changlian Ma, Haian Huang, Jianfei Gao, et~al.
\newblock {InternVideo2.5: Empowering Video MLLMs with Long and Rich Context
  Modeling}.
\newblock \emph{arXiv preprint arXiv:2501.12386}, 2025{\natexlab{b}}.

\bibitem[Wang et~al.(2025{\natexlab{c}})Wang, Meng, Wang, Liang, Wei, Zhang,
  and Zhao]{mmduet}
Yueqian Wang, Xiaojun Meng, Yuxuan Wang, Jianxin Liang, Jiansheng Wei, Huishuai
  Zhang, and Dongyan Zhao.
\newblock {VideoLLM Knows When to Speak: Enhancing Time-Sensitive Video
  Comprehension with Video-Text Duet Interaction Format}.
\newblock In \emph{Findings of the Association for Computational Linguistics:
  EMNLP 2025}, 2025{\natexlab{c}}.

\bibitem[Wang et~al.(2025{\natexlab{d}})Wang, Meng, Wang, Zhang, and
  Zhao]{proactivevideoqa}
Yueqian Wang, Xiaojun Meng, Yifan Wang, Huishuai Zhang, and Dongyan Zhao.
\newblock {ProactiveVideoQA: A Comprehensive Benchmark Evaluating Proactive
  Interactions in Video Large Language Models}.
\newblock \emph{arXiv preprint arXiv:2507.09313}, 2025{\natexlab{d}}.

\bibitem[Wang et~al.(2025{\natexlab{e}})Wang, Wang, Chen, Wu, Zhao, and
  Zheng]{omnimmi}
Yuxuan Wang, Yueqian Wang, Bo Chen, Tong Wu, Dongyan Zhao, and Zilong Zheng.
\newblock {OmniMMI: A Comprehensive Multi-modal Interaction Benchmark in
  Streaming Video Contexts}.
\newblock In \emph{Proceedings of the IEEE/CVF Conference on Computer Vision
  and Pattern Recognition}, pages 18925--18935, 2025{\natexlab{e}}.

\bibitem[Wang et~al.(2026{\natexlab{b}})Wang, Liu, Wang, Xu, Wan, Zhang, and
  Zhao]{mmduet2}
Yueqian Wang, Songxiang Liu, Disong Wang, Nuo Xu, Guanglu Wan, Huishuai Zhang,
  and Dongyan Zhao.
\newblock {MMDuet2: Enhancing Proactive Interaction of Video MLLMs with
  Multi-Turn Reinforcement Learning}.
\newblock In \emph{International Conference on Learning Representations},
  2026{\natexlab{b}}.

\bibitem[Wen et~al.(2026)Wen, Yang, Ke, Huang, Liao, Wang, Liu, and
  Zhang]{evostreaming}
Zichen Wen, Boxue Yang, Junlong Ke, Jiajie Huang, Chenfei Liao, Junxi Wang,
  Xuyang Liu, and Linfeng Zhang.
\newblock {EvoStreaming: Your Offline Video Model Is a Natively Streaming
  Assistant}.
\newblock \emph{arXiv preprint arXiv:2605.10343}, 2026.

\bibitem[Wu et~al.(2024)Wu, Li, Chen, and Li]{longvideobench}
Haoning Wu, Dongxu Li, Bei Chen, and Junnan Li.
\newblock {LongVideoBench: A Benchmark for Long-context Interleaved
  Video-Language Understanding}.
\newblock In \emph{Advances in Neural Information Processing Systems}, 2024.

\bibitem[Xia et~al.(2026)Xia, Chen, Zhang, Sun, and Zhou]{streamo}
Jiaer Xia, Peixian Chen, Mengdan Zhang, Xing Sun, and Kaiyang Zhou.
\newblock {Streaming Video Instruction Tuning}.
\newblock In \emph{Proceedings of the IEEE/CVF Conference on Computer Vision
  and Pattern Recognition}, 2026.

\bibitem[Xu et~al.(2026)Xu, Xiao, Chen, He, Lu, and Han]{streamingvlm}
Ruyi Xu, Guangxuan Xiao, Yukang Chen, Liuning He, Yao Lu, and Song Han.
\newblock {StreamingVLM: Real-Time Understanding for Infinite Video Streams}.
\newblock In \emph{International Conference on Learning Representations}, 2026.

\bibitem[Yan et~al.(2026)Yan, Dai, Ran, Li, Lin, Jin, Xie, Liao, and
  Lian]{proactvl}
Weicai Yan, Yuhong Dai, Qi Ran, Haodong Li, Wang Lin, Tao Jin, Xing Xie, Hao
  Liao, and Jianxun Lian.
\newblock {Proact-VL: A Proactive VideoLLM for Real-Time AI Companions}.
\newblock \emph{arXiv preprint arXiv:2603.03447}, 2026.

\bibitem[Yang et~al.(2026)Yang, Zhang, Jia, Guo, Shen, Zhang, Zhang, Wang, Li,
  Zhang, et~al.]{mage}
Senqiao Yang, Kaichen Zhang, Zhaoyang Jia, Jinghao Guo, Yifei Shen, Xinjie
  Zhang, Xiaoyi Zhang, Haoqing Wang, Xiao Li, Peng Zhang, et~al.
\newblock {Mage-VL: An Efficient Codec-Native Streaming Multimodal Foundation
  Model}.
\newblock \emph{arXiv preprint arXiv:2607.24904}, 2026.

\bibitem[Yang et~al.(2025)Yang, Zhang, Hu, Wang, Qian, Wen, Yang, Gao, Dong,
  and Xu]{livestar}
Zhenyu Yang, Kairui Zhang, Yuhang Hu, Bing Wang, Shengsheng Qian, Bin Wen, Fan
  Yang, Tingting Gao, Weiming Dong, and Changsheng Xu.
\newblock {LiveStar: Live Streaming Assistant for Real-World Online Video
  Understanding}.
\newblock In \emph{Advances in Neural Information Processing Systems}, 2025.

\bibitem[Yao et~al.(2026)Yao, Zhou, Yang, Qin, Zeng, Li, Hou, Liang, Wang, Tuo,
  Zhang, Zhu, Cao, Ye, Xie, Gu, Huang, Si, Duan, and Wang]{joyai}
Dingyu Yao, Junhao Zhou, Chenxu Yang, Chuanyu Qin, Xiangyu Zeng, Yifei Li,
  Haowen Hou, Zheming Liang, Congcong Wang, Kaiwen Tuo, Jun Zhang, Yuhan Zhu,
  Yuhang Cao, Shenglong Ye, Shuai Xie, Shuhuan Gu, Haoyang Huang, Qingyi Si,
  Nan Duan, and Jiaqi Wang.
\newblock {JoyAI-VL-Interaction: Real-Time Vision-Language Interaction
  Intelligence}.
\newblock \emph{arXiv preprint arXiv:2606.14777}, 2026.

\bibitem[Yao et~al.(2025)Yao, Li, Wei, Li, Ren, Liu, Ouyang, Wang, Li, Li,
  et~al.]{timechatonline}
Linli Yao, Yicheng Li, Yuancheng Wei, Lei Li, Shuhuai Ren, Yuanxin Liu, Kun
  Ouyang, Lean Wang, Shicheng Li, Sida Li, et~al.
\newblock {TimeChat-Online: 80\% Visual Tokens are Naturally Redundant in
  Streaming Videos}.
\newblock In \emph{Proceedings of the ACM International Conference on
  Multimedia}, 2025.

\bibitem[Yao et~al.(2023)Yao, Zhao, Yu, Du, Shafran, Narasimhan, and
  Cao]{react}
Shunyu Yao, Jeffrey Zhao, Dian Yu, Nan Du, Izhak Shafran, Karthik Narasimhan,
  and Yuan Cao.
\newblock {ReAct: Synergizing Reasoning and Acting in Language Models}.
\newblock In \emph{International Conference on Learning Representations}, 2023.

\bibitem[Zeng et~al.(2026)Zeng, Yang, Zhang, Zhu, Li, Si, Yao, Ma, Chen, Chen,
  Shi, Zhou, Li, Zhang, Qin, Yang, Yu, Ouyang, Shao, Wei, Zhou, Gao, Wang, and
  Wang]{onestreamer}
Xiangyu Zeng, Yuandong Yang, Zhiqiu Zhang, Yuhan Zhu, Xinhao Li, Qingyi Si,
  Dingyu Yao, Changlian Ma, Haoran Chen, Xinyu Chen, Yansong Shi, Junhao Zhou,
  Yifei Li, Jun Zhang, Chuanyu Qin, Chenxu Yang, Xinlei Yu, Kun Ouyang, Yuchen
  Shao, Qianshan Wei, Changhai Zhou, Jun Gao, Jiaqi Wang, and Limin Wang.
\newblock {OneStreamer: Unifying Perception, Memory, and Proactive Response in
  Streaming Video Interaction}.
\newblock \emph{arXiv preprint arXiv:2610.01762}, 2026.

\bibitem[Zhang et~al.(2025)Zhang, Wang, Tang, Liu, Feng, and Jin]{flashvstream}
Haoji Zhang, Yiqin Wang, Yansong Tang, Yong Liu, Jiashi Feng, and Xiaojie Jin.
\newblock {Flash-VStream: Efficient Real-Time Understanding for Long Video
  Streams}.
\newblock In \emph{Proceedings of the IEEE/CVF International Conference on
  Computer Vision}, 2025.

\bibitem[Zhang et~al.(2024{\natexlab{a}})Zhang, Zhang, Li, Zeng, Yang, Zhang,
  Wang, Tan, Li, and Liu]{longva}
Peiyuan Zhang, Kaichen Zhang, Bo Li, Guangtao Zeng, Jingkang Yang, Yuanhan
  Zhang, Ziyue Wang, Haoran Tan, Chunyuan Li, and Ziwei Liu.
\newblock {Long Context Transfer from Language to Vision}.
\newblock \emph{arXiv preprint arXiv:2406.16852}, 2024{\natexlab{a}}.

\bibitem[Zhang et~al.(2024{\natexlab{b}})Zhang, Wu, Li, Li, Ma, Liu, and
  Li]{llavavideo}
Yuanhan Zhang, Jinming Wu, Wei Li, Bo Li, Zejun Ma, Ziwei Liu, and Chunyuan Li.
\newblock {LLaVA-Video: Video Instruction Tuning With Synthetic Data}.
\newblock \emph{arXiv preprint arXiv:2410.02713}, 2024{\natexlab{b}}.

\bibitem[Zhu et~al.(2025)Zhu, Wang, Chen, Liu, Ye, Gu, Tian, Duan, Su, Shao,
  et~al.]{internvl3}
Jinguo Zhu, Weiyun Wang, Zhe Chen, Zhaoyang Liu, Shenglong Ye, Lixin Gu, Hao
  Tian, Yuchen Duan, Weijie Su, Jie Shao, et~al.
\newblock {InternVL3: Exploring Advanced Training and Test-Time Recipes for
  Open-Source Multimodal Models}.
\newblock \emph{arXiv preprint arXiv:2504.10479}, 2025.

\end{thebibliography}
}

\clearpage
\onecolumn
\appendix
\etocdepthtag.toc{mtappendix}
\raggedbottom
\begin{center}
{\Large\bfseries Appendix}
\end{center}
\vspace{2pt}
\noindent The appendix first gives the evaluation details and full results, then extends the analysis of general VLMs, describes the protocols and prompts of \livecue, and reports ablations, the comparison with IPI-Agent, and efficiency.

\etocsettagdepth{mtmain}{none}
\etocsettagdepth{mtappendix}{subsection}
\etocsettocstyle{\vspace{10pt}\noindent{\large\bfseries Contents}\par\vspace{4pt}\hrule\vspace{6pt}}{\vspace{4pt}\hrule\vspace{10pt}}
\tableofcontents

\section{Benchmarks and Evaluation}
\label{sec:app:benchmarks}

\begin{table}[H]
\centering
\small
\begin{tabular}{@{}p{0.42\linewidth}rp{0.39\linewidth}@{}}
\toprule
Benchmark & $N$ & Reported measure \\
\midrule
ProactiveVQA~\cite{proactivevideoqa} & 1,427 & PAUC, $\omega=0.5$ \\
OVO-Timing~\cite{ovobench} & 172 & F1@2s \\
OmniMMI-PA~\cite{omnimmi} & 200 & First-trigger accuracy \\
ViSpeak~\cite{vispeak} & 800 & Task macro score \\
InteractionBench~\cite{interactionbench} & 1,060 & Official Overall \\
OmniStar-RNG~\cite{livestar} & 200 & RNG, Eq.~\ref{eq:rng-balance} \\
\bottomrule
\end{tabular}
\caption{Benchmarks and native streaming evaluation measures.}
\label{tab:benchmarks}
\end{table}

\paragraph{Metrics.}
Table~\ref{tab:benchmarks} lists the score reported for each benchmark in Table~\ref{tab:main-results}. Overall averages the six benchmark scores. The native metrics are reported below.

\paragraph{Balanced OmniStar-RNG score.}
OmniStar-RNG reports five video-macro metrics for real-time narration: semantic correctness $S\in[0,10]$, fluency $F\in[0,10]$, temporal coverage $C\in[0,1]$, temporal deviation $D$ in seconds, and temporal redundancy $R$. They pull in opposite directions, so any single one can be improved at the expense of the others: a silent system has no redundancy and no deviation, while a system that narrates at every step approaches full coverage with heavy redundancy. We therefore combine all five into
\begin{equation}
 R_{\mathrm{bal}} = 100\left[\frac{(S/10)(F/10)\,C}{(1+D/\tau)(1+R)}\right]^{1/5},\quad \tau=1\,\mathrm{s}.
 \label{eq:rng-balance}
\end{equation}
Each term is normalized to $[0,1]$, with deviation and redundancy entering as $1/(1+\cdot)$, and the geometric mean requires all five aspects to hold at once: a collapse in any one, such as zero coverage, drives the score toward zero, and no term can compensate for another. In Table~\ref{tab:app:rng}, AURA, which stays silent, receives 0.0, and JoyAI-VL-Interaction, which covers 99.3\% of the timeline with a redundancy of 2.16, is credited for its coverage and content but penalized for its redundancy.

\paragraph{Stability of the RNG score.}
Table~\ref{tab:app:rng-stability} recomputes the score from the same native metrics under alternative choices. With Qwen3.8-27B, \livecue obtains the highest score for every deviation scale from 0.5\,s to 5\,s and when the five terms are averaged arithmetically; the ordering of \livecue against Direct and the agent frameworks is unchanged on all three VLMs. Leaving out any one of semantic correctness, fluency, coverage, or deviation also keeps \livecue on Qwen3.8-27B first (52.4--61.7). Leaving out redundancy instead favors systems that narrate almost continuously: JoyAI-system on Qwen3.8-27B, which covers 93.2\% of the timeline with a redundancy of 3.15, then reaches 64.4 against 63.2, which illustrates why redundancy must be constrained jointly with coverage. Finally, averaging the other five benchmarks without OmniStar-RNG, \livecue on Qwen3.8-27B obtains 53.2, ahead of every streaming-trained model and agent framework.
\begin{table}[H]
\centering
\small
\setlength{\tabcolsep}{5pt}
\renewcommand{\arraystretch}{1.12}
\begin{tabular}{l|cccc|c|c}
\toprule
& \multicolumn{4}{c|}{Balanced RNG, $\tau$ (s)} & \multirow{2}{*}{Arith.} & \multirow{2}{*}{\shortstack{Overall\\w/o RNG}} \\
\textbf{Method} & 0.5 & 1 & 2 & 5 & & \\
\midrule
\multicolumn{7}{l}{\textit{Streaming-trained models}} \\
Mage-VL~\cite{mage} & 13.8 & 15.3 & 16.7 & 18.1 & 26.7 & 19.2 \\
JoyAI-VL-Interaction~\cite{joyai} & 52.7 & 54.8 & 56.1 & 56.9 & 59.8 & 28.1 \\
AURA~\cite{aura} & 0.0 & 0.0 & 0.0 & 0.0 & 14.6 & 27.9 \\
MOSS-VL-Realtime~\cite{moss} & 22.3 & 24.8 & 27.1 & 29.4 & 27.7 & 44.6 \\
VideoChat3~\cite{videochat3} & 31.3 & 34.5 & 37.2 & 39.7 & 36.5 & 27.4 \\
\midrule
\textbf{Qwen3.5-4B} & 26.7 & 29.3 & 31.6 & 33.6 & 31.4 & 23.0 \\
\quad + JoyAI system & 18.8 & 21.1 & 23.1 & 25.3 & 25.3 & 27.5 \\
\quad + IPI-Agent & 25.2 & 28.0 & 30.4 & 32.7 & 33.8 & 11.8 \\
\rowcolor{ocrow}
\quad + \livecue & 52.1 & 56.0 & 58.8 & 61.0 & 57.9 & 40.5 \\
\midrule
\textbf{Qwen3.6-35B-A3B} & 32.3 & 35.4 & 38.0 & 40.3 & 36.6 & 20.4 \\
\quad + JoyAI system & 47.5 & 50.6 & 52.7 & 54.3 & 53.6 & 23.0 \\
\quad + IPI-Agent & 33.2 & 37.0 & 40.6 & 44.3 & 40.1 & 23.0 \\
\rowcolor{ocrow}
\quad + \livecue & 50.5 & 53.1 & 54.8 & 56.0 & 55.4 & 44.8 \\
\midrule
\textbf{Qwen3.8-27B} & 19.5 & 21.8 & 23.9 & 26.0 & 24.0 & 42.2 \\
\quad + JoyAI system & 51.9 & 52.9 & 53.5 & 53.8 & 59.5 & 34.7 \\
\quad + IPI-Agent & 27.8 & 31.2 & 34.4 & 37.8 & 35.6 & 26.4 \\
\rowcolor{ocrow}
\quad + \livecue & \textbf{55.4} & \textbf{59.1} & \textbf{61.7} & \textbf{63.6} & \textbf{61.3} & \textbf{53.2} \\
\bottomrule
\end{tabular}
\caption{\textbf{Stability of the OmniStar-RNG score.} The balanced score of Eq.~\ref{eq:rng-balance} recomputed from the native metrics of Table~\ref{tab:app:rng} with different deviation scales $\tau$ ($\tau=1$\,s is the reported score), with the five normalized terms averaged arithmetically instead of geometrically (Arith.), and the Overall of Table~\ref{tab:main-results} computed without OmniStar-RNG. The first row of each VLM group is Direct. Best values are in bold.}
\label{tab:app:rng-stability}
\end{table}

\paragraph{Evaluation settings.}
Video observations arrive causally, and proactive outputs are evaluated at their video-time positions. Native proactive models use their benchmark-specific streaming protocols. All systems are run by us and scored with the same evaluators and content judge. The general-purpose model and framework configurations follow Sec.~\ref{sec:exp:setup}. Content scoring retains the benchmark-specific rubrics. The tables use percentage scales for PAUC, timing accuracy, InteractionBench Total, and benchmark-level Overall scores, while preserving the native scales of the ViSpeak text and RNG component measures.

\paragraph{Choice of content judge.}
The content judge, Qwen3.5-35B-A3B, comes from the same model family as the backbones of \livecue. To test whether the comparison depends on this choice, we replace it with DeepSeek-V4.1-Flash~\cite{deepseekv41}, run in non-thinking mode at temperature 0, and rescore \livecue on Qwen3.8-27B and MOSS-VL-Realtime, the strongest streaming-trained model in Table~\ref{tab:main-results}. As Table~\ref{tab:app:judge} shows, absolute content scores shift with the judge, falling on ProactiveVideoQA and ViSpeak and rising on OmniStar-RNG, while \livecue stays ahead on all four benchmarks with content judgments. Its Overall margin over MOSS-VL-Realtime grows from 12.9 to 13.7 points.
\begin{table}[H]
\centering
\small
\setlength{\tabcolsep}{4pt}
\begin{tabular}{@{}llccccccc@{}}
\toprule
Judge & System & PVQA & OVO & PA & ViSpeak & IB & RNG & Overall \\
\midrule
Qwen3.5-35B-A3B & MOSS-VL-Realtime & 41.1 & 38.7 & 57.5 & 48.4 & 37.3 & 24.8 & 41.3 \\
 & \livecue (Qwen3.8-27B) & \textbf{47.9} & \textbf{41.1} & \textbf{61.5} & \textbf{63.1} & \textbf{52.6} & \textbf{59.1} & \textbf{54.2} \\
\midrule
DeepSeek-V4.1-Flash & MOSS-VL-Realtime & 34.7 & 38.7 & 57.5 & 45.6 & 37.1 & 25.3 & 39.8 \\
 & \livecue (Qwen3.8-27B) & \textbf{43.3} & \textbf{41.1} & \textbf{61.5} & \textbf{61.1} & \textbf{52.6} & \textbf{61.6} & \textbf{53.5} \\
\bottomrule
\end{tabular}
\caption{\textbf{Replacing the content judge.} Predictions, evaluated items, rubric prompts, and metric definitions are fixed; only the content judge changes. OVO-Timing and OmniMMI-PA do not use the content judge and keep their values; on OmniStar-RNG, semantic correctness and fluency are rejudged and the temporal metrics are kept.}
\label{tab:app:judge}
\end{table}

\FloatBarrier

\section{Per-benchmark Results}
\label{sec:app:perbench}

Tables~\ref{tab:app:pvqa}--\ref{tab:app:rng} expand the benchmark scores of Table~\ref{tab:main-results} into each benchmark's native task-level metrics, with the same model and framework configurations. For each general-purpose VLM, the first row reports the VLM used directly (Direct), followed by the two agent frameworks and \livecue built on it.

\begin{table}[H]
\centering
\small
\setlength{\tabcolsep}{2pt}
\renewcommand{\arraystretch}{1.12}
\begin{adjustbox}{max width=\linewidth}
\begin{tabular}{lc|ccc|ccc|ccc|ccc|c}
\toprule
\multirow{2}{*}{\textbf{Method}} & \multirow{2}{*}{\textbf{Size}} & \multicolumn{3}{c|}{\textbf{WEB}} & \multicolumn{3}{c|}{\textbf{EGO}} & \multicolumn{3}{c|}{\textbf{TV}} & \multicolumn{3}{c|}{\textbf{VAD}} & \multirow{2}{*}{\textbf{Avg.}} \\
\cmidrule(lr){3-5}\cmidrule(lr){6-8}\cmidrule(lr){9-11}\cmidrule(lr){12-14}
 & & 0.0 & 0.5 & 1.0 & 0.0 & 0.5 & 1.0 & 0.0 & 0.5 & 1.0 & 0.0 & 0.5 & 1.0 & \\
\midrule
\multicolumn{15}{l}{\textit{Streaming-trained models}} \\
Mage-VL~\cite{mage} & 4B & 26.8 & 27.2 & 27.5 & 24.9 & 24.9 & 24.8 & 25.6 & 25.6 & 25.7 & 25.0 & 25.1 & 25.2 & 25.7 \\
JoyAI-VL-Interaction~\cite{joyai} & 8B & 34.5 & 38.5 & 42.5 & 28.8 & 30.1 & 31.4 & 31.2 & 32.4 & 33.7 & 27.1 & 28.1 & 29.1 & 32.3 \\
AURA~\cite{aura} & 8B & 34.6 & 39.7 & 44.8 & 45.6 & 50.7 & 55.9 & 31.8 & 34.7 & 37.6 & 27.8 & 30.8 & 33.8 & 39.0 \\
MOSS-VL-Realtime~\cite{moss} & 11B & 37.0 & 42.0 & 47.0 & 38.1 & 40.7 & 43.2 & 41.9 & 50.2 & 58.5 & 29.5 & 31.7 & 33.9 & 41.1 \\
VideoChat3~\cite{videochat3} & 4B & 30.7 & 32.3 & 33.8 & 25.7 & 26.0 & 26.4 & 33.4 & 38.1 & 42.8 & 25.4 & 25.8 & 26.2 & 30.6 \\
\midrule
\textbf{Qwen3.5-4B} & 4B & 41.3 & 45.4 & 49.5 & 33.8 & 34.6 & 35.5 & 37.4 & 41.2 & 44.9 & 26.2 & 27.0 & 27.8 & 37.1 \\
\quad + JoyAI system & 4B & 30.1 & 31.7 & 33.3 & 26.8 & 27.5 & 28.3 & 27.6 & 28.4 & 29.3 & 27.8 & 29.0 & 30.1 & 29.2 \\
\quad + IPI-Agent & 4B & 19.2 & 19.2 & 19.2 & 21.1 & 20.9 & 20.7 & 21.9 & 22.0 & 22.0 & 13.5 & 13.2 & 12.9 & 18.8 \\
\rowcolor{ocrow}
\quad + \livecue & 4B & 44.3 & 49.3 & 54.4 & 38.5 & 41.3 & 44.2 & 39.0 & 44.1 & 49.2 & 32.3 & 34.9 & 37.4 & 42.4 \\
\midrule
\textbf{Qwen3.6-35B-A3B} & 35B-A3B & 45.5 & 49.3 & 53.1 & 27.7 & 27.9 & 28.0 & 37.7 & 41.0 & 44.3 & 26.6 & 26.9 & 27.1 & 36.3 \\
\quad + JoyAI system & 35B-A3B & 49.6 & 54.6 & 59.5 & 36.3 & 37.6 & 38.8 & 52.2 & 56.9 & 61.6 & 29.7 & 32.5 & 35.3 & 45.4 \\
\quad + IPI-Agent & 35B-A3B & 18.8 & 18.8 & 18.8 & 25.0 & 25.0 & 25.0 & 21.8 & 21.8 & 21.8 & 30.5 & 31.7 & 32.9 & 24.3 \\
\rowcolor{ocrow}
\quad + \livecue & 35B-A3B & 49.2 & 55.2 & 61.1 & 39.0 & 42.6 & 46.2 & 46.6 & 53.1 & 59.7 & 35.9 & 38.5 & 41.1 & 47.3 \\
\midrule
\textbf{Qwen3.8-27B} & 27B & 44.2 & 48.1 & 51.9 & 27.2 & 27.9 & 28.5 & 51.9 & 59.6 & 67.3 & 30.6 & 32.2 & 33.9 & 41.9 \\
\quad + JoyAI system & 27B & 56.2 & 62.0 & 67.7 & 32.6 & 33.5 & 34.5 & 56.8 & 62.5 & 68.3 & 28.1 & 30.9 & 33.6 & 47.2 \\
\quad + IPI-Agent & 27B & 26.9 & 26.9 & 26.9 & 25.1 & 25.1 & 25.1 & 21.5 & 21.5 & 21.5 & 25.0 & 25.0 & 25.0 & 24.6 \\
\rowcolor{ocrow}
\quad + \livecue & 27B & 47.6 & 52.9 & 58.1 & 39.7 & 43.6 & 47.4 & 50.0 & 58.3 & 66.6 & 34.2 & 36.8 & 39.5 & 47.9 \\
\bottomrule
\end{tabular}
\end{adjustbox}
\caption{\textbf{Detailed evaluation results on ProactiveVideoQA.} The first row of each VLM group is the VLM used directly (Direct). PAUC scores at $\omega=0.0$, $0.5$, and $1.0$. Avg. is the unweighted four-task mean at $\omega=0.5$.}
\label{tab:app:pvqa}
\end{table}

\begin{table}[H]
\centering
\small
\setlength{\tabcolsep}{2pt}
\renewcommand{\arraystretch}{1.12}
\begin{adjustbox}{max width=\linewidth}
\begin{tabular}{lc|ccc|ccc|ccc|c}
\toprule
\multirow{2}{*}{\textbf{Method}} & \multirow{2}{*}{\textbf{Size}} & \multicolumn{3}{c|}{\textbf{CRR}} & \multicolumn{3}{c|}{\textbf{SSR}} & \multicolumn{3}{c|}{\textbf{REC}} & \multirow{2}{*}{\textbf{Avg. F1}} \\
\cmidrule(lr){3-5}\cmidrule(lr){6-8}\cmidrule(lr){9-11}
 & & R & P & F1 & R & P & F1 & R & P & F1 & \\
\midrule
\multicolumn{12}{l}{\textit{Streaming-trained models}} \\
Mage-VL~\cite{mage} & 4B & 41.7 & 8.7 & 14.5 & 54.2 & 11.2 & 18.6 & 38.3 & 27.3 & 31.9 & 21.7 \\
JoyAI-VL-Interaction~\cite{joyai} & 8B & 58.3 & 8.5 & 14.9 & 44.5 & 7.1 & 12.2 & 67.0 & 25.2 & 36.7 & 21.3 \\
AURA~\cite{aura} & 8B & 93.8 & 3.4 & 6.5 & 15.1 & 5.6 & 8.2 & 41.9 & 13.5 & 20.5 & 11.7 \\
MOSS-VL-Realtime~\cite{moss} & 11B & 41.7 & 21.2 & 28.1 & 46.4 & 21.6 & 29.4 & 54.2 & 63.2 & 58.4 & 38.7 \\
VideoChat3~\cite{videochat3} & 4B & 35.4 & 28.1 & 31.4 & 43.7 & 16.1 & 23.5 & 45.6 & 46.1 & 45.8 & 33.6 \\
\midrule
\textbf{Qwen3.5-4B} & 4B & 83.3 & 2.4 & 4.7 & 30.9 & 7.6 & 12.2 & 67.9 & 13.1 & 22.0 & 13.0 \\
\quad + JoyAI system & 4B & 16.7 & 14.2 & 15.4 & 19.6 & 9.8 & 13.1 & 28.0 & 33.5 & 30.5 & 19.6 \\
\quad + IPI-Agent & 4B & 6.3 & 1.6 & 2.5 & 18.5 & 6.2 & 9.3 & 11.5 & 22.6 & 15.2 & 9.0 \\
\rowcolor{ocrow}
\quad + \livecue & 4B & 31.3 & 18.8 & 23.5 & 67.6 & 19.0 & 29.6 & 20.0 & 25.3 & 22.4 & 25.2 \\
\midrule
\textbf{Qwen3.6-35B-A3B} & 35B-A3B & 75.0 & 2.1 & 4.2 & 18.9 & 3.2 & 5.4 & 41.9 & 21.1 & 28.1 & 12.6 \\
\quad + JoyAI system & 35B-A3B & 70.8 & 6.3 & 11.6 & 81.3 & 3.9 & 7.4 & 72.7 & 18.7 & 29.8 & 16.3 \\
\quad + IPI-Agent & 35B-A3B & 6.3 & 0.8 & 1.5 & 7.8 & 3.4 & 4.7 & 40.9 & 22.3 & 28.8 & 11.7 \\
\rowcolor{ocrow}
\quad + \livecue & 35B-A3B & 68.8 & 18.7 & 29.4 & 82.3 & 16.4 & 27.4 & 32.0 & 53.1 & 40.0 & 32.3 \\
\midrule
\textbf{Qwen3.8-27B} & 27B & 37.5 & 21.6 & 27.4 & 4.6 & 4.5 & 4.5 & 22.1 & 32.5 & 26.3 & 19.4 \\
\quad + JoyAI system & 27B & 93.8 & 3.0 & 5.7 & 56.7 & 2.5 & 4.8 & 86.4 & 13.4 & 23.2 & 11.2 \\
\quad + IPI-Agent & 27B & 0.0 & 0.0 & 0.0 & 9.0 & 13.2 & 10.7 & 23.5 & 35.1 & 28.2 & 13.0 \\
\rowcolor{ocrow}
\quad + \livecue & 27B & 47.9 & 31.3 & 37.9 & 63.7 & 22.2 & 32.9 & 49.6 & 55.8 & 52.5 & 41.1 \\
\bottomrule
\end{tabular}
\end{adjustbox}
\caption{\textbf{Detailed evaluation results on OVO-Timing.} The first row of each VLM group is the VLM used directly (Direct). Recall (R), precision (P), and F1 use a two-second tolerance. Avg. F1 is the unweighted mean over CRR, SSR, and REC.}
\label{tab:app:ovotiming}
\end{table}

\begin{table}[H]
\centering
\small
\setlength{\tabcolsep}{2pt}
\renewcommand{\arraystretch}{1.12}
\begin{adjustbox}{max width=\linewidth}
\begin{tabular}{lc|ccccccc|ccccccc|c}
\toprule
\multirow{2}{*}{\textbf{Method}} & \multirow{2}{*}{\textbf{Size}} & \multicolumn{7}{c|}{\textbf{Time Accuracy (\%)}} & \multicolumn{7}{c|}{\textbf{Text Score (0--5)}} & \multirow{2}{*}{\textbf{Overall}} \\
\cmidrule(lr){3-9}\cmidrule(lr){10-16}
 & & AW & VI & HR & VW & VT & GU & \textbf{All} & AW & VI & HR & VW & VT & GU & \textbf{All} & \\
\midrule
\multicolumn{17}{l}{\textit{Streaming-trained models}} \\
Mage-VL~\cite{mage} & 4B & 55.00 & 65.00 & 82.00 & 55.00 & 26.00 & 56.00 & 56.50 & 0.88 & 0.00 & 0.17 & 3.91 & 0.00 & 1.51 & 1.08 & 12.07 \\
JoyAI-VL-Interaction~\cite{joyai} & 8B & 19.00 & 16.00 & 60.00 & 97.00 & 57.00 & 92.50 & 56.92 & 2.16 & 4.88 & 0.97 & 5.00 & 4.42 & 2.92 & 3.39 & 39.48 \\
AURA~\cite{aura} & 8B & 34.00 & 22.00 & 2.00 & 0.00 & 78.00 & 4.00 & 23.33 & 1.46 & 1.27 & 1.00 & 0.00 & 4.79 & 3.63 & 2.02 & 15.60 \\
MOSS-VL-Realtime~\cite{moss} & 11B & 54.00 & 66.00 & 39.00 & 95.00 & 87.00 & 99.00 & 73.33 & 2.61 & 2.83 & 0.97 & 5.00 & 4.89 & 1.87 & 3.03 & 48.38 \\
VideoChat3~\cite{videochat3} & 4B & 13.50 & 1.00 & 1.00 & 58.00 & 26.00 & 71.50 & 28.50 & 0.85 & 3.00 & 2.00 & 5.00 & 4.73 & 1.66 & 2.87 & 18.28 \\
\midrule
\textbf{Qwen3.5-4B} & 4B & 41.00 & 15.00 & 56.00 & 92.00 & 75.00 & 77.00 & 59.33 & 1.20 & 3.47 & 0.75 & 5.00 & 0.93 & 2.40 & 2.29 & 28.60 \\
\quad + JoyAI system & 4B & 36.00 & 38.00 & 40.00 & 90.00 & 91.00 & 83.50 & 63.08 & 1.92 & 3.21 & 1.05 & 5.00 & 4.70 & 2.56 & 3.07 & 44.15 \\
\quad + IPI-Agent & 4B & 0.00 & 45.00 & 48.00 & 79.00 & 55.00 & 99.50 & 54.42 & 0.00 & 0.11 & 0.33 & 0.00 & 0.93 & 0.54 & 0.32 & 4.20 \\
\rowcolor{ocrow}
\quad + \livecue & 4B & 47.00 & 40.00 & 42.00 & 98.00 & 96.00 & 97.00 & 70.00 & 2.28 & 5.00 & 1.29 & 5.00 & 5.00 & 2.25 & 3.47 & 51.63 \\
\midrule
\textbf{Qwen3.6-35B-A3B} & 35B-A3B & 11.50 & 6.00 & 18.00 & 27.00 & 10.00 & 11.50 & 14.00 & 2.57 & 4.17 & 2.33 & 5.00 & 5.00 & 1.22 & 3.38 & 9.85 \\
\quad + JoyAI system & 35B-A3B & 17.50 & 19.00 & 45.00 & 83.00 & 24.00 & 100.00 & 48.08 & 1.83 & 4.21 & 1.27 & 4.65 & 4.92 & 2.30 & 3.19 & 30.08 \\
\quad + IPI-Agent & 35B-A3B & 33.00 & 37.00 & 49.00 & 99.00 & 64.00 & 100.00 & 63.67 & 2.24 & 4.97 & 1.06 & 5.00 & 5.00 & 2.35 & 3.44 & 45.33 \\
\rowcolor{ocrow}
\quad + \livecue & 35B-A3B & 48.00 & 69.00 & 51.00 & 99.00 & 98.00 & 100.00 & 77.50 & 2.60 & 4.97 & 1.29 & 5.00 & 5.00 & 2.47 & 3.56 & 58.85 \\
\midrule
\textbf{Qwen3.8-27B} & 27B & 50.50 & 61.00 & 35.00 & 97.00 & 97.00 & 75.50 & 69.33 & 4.14 & 4.90 & 1.74 & 5.00 & 5.00 & 3.11 & 3.98 & 59.13 \\
\quad + JoyAI system & 27B & 26.00 & 74.00 & 58.00 & 99.00 & 97.00 & 99.00 & 75.50 & 2.96 & 4.58 & 1.36 & 5.00 & 4.99 & 3.14 & 3.67 & 59.50 \\
\quad + IPI-Agent & 27B & 42.50 & 39.00 & 54.00 & 99.00 & 67.00 & 96.00 & 66.25 & 2.95 & 3.92 & 0.91 & 3.97 & 4.87 & 2.56 & 3.20 & 43.08 \\
\rowcolor{ocrow}
\quad + \livecue & 27B & 50.50 & 67.00 & 41.00 & 99.00 & 98.00 & 98.50 & 75.67 & 3.78 & 5.00 & 1.98 & 5.00 & 5.00 & 3.06 & 3.97 & 63.12 \\
\bottomrule
\end{tabular}
\end{adjustbox}
\caption{\textbf{Detailed evaluation results on ViSpeak-Bench.} The first row of each VLM group is the VLM used directly (Direct). Tasks: anomaly warning (AW), visual interruption (VI), humor reaction (HR), visual wake-up (VW), visual termination (VT), and gesture understanding (GU). Time All and Text All are unweighted means over the six tasks. Text scores are conditioned on correct timing. Overall is the timing-gated text macro score on a 0--100 scale.}
\label{tab:app:vispeak}
\end{table}

\begin{table}[H]
\centering
\small
\setlength{\tabcolsep}{2pt}
\renewcommand{\arraystretch}{1.12}
\begin{adjustbox}{max width=\linewidth}
\begin{tabular}{lc|cccccccc|cccc|c}
\toprule
\multirow{2}{*}{\textbf{Method}} & \multirow{2}{*}{\textbf{Size}} & \multicolumn{8}{c|}{\textbf{Capability}} & \multicolumn{4}{c|}{\textbf{Response Requirement}} & \multirow{2}{*}{\textbf{Overall}} \\
\cmidrule(lr){3-10}\cmidrule(lr){11-14}
 & & IVQA & LVM & CIR & TOA & PTR & CST & BRC & LCG & A & B & C & B/C & \\
\midrule
\multicolumn{15}{l}{\textit{Streaming-trained models}} \\
Mage-VL~\cite{mage} & 4B & 40.1 & 41.3 & 28.6 & 31.3 & 33.0 & 5.5 & 100.0 & 41.0 & 40.8 & 33.1 & 21.6 & 26.9 & 31.0 \\
JoyAI-VL-Interaction~\cite{joyai} & 8B & 34.8 & 60.9 & 7.1 & 22.0 & 37.7 & 4.4 & 100.0 & 23.4 & 47.8 & 30.5 & 19.0 & 24.2 & 31.2 \\
AURA~\cite{aura} & 8B & 59.1 & 14.2 & 0.0 & 24.8 & 39.0 & 11.2 & 100.0 & 15.1 & 37.0 & 32.4 & 20.1 & 25.7 & 29.1 \\
MOSS-VL-Realtime~\cite{moss} & 11B & 73.2 & 85.4 & 7.1 & 19.7 & 23.2 & 16.4 & 0.0 & 15.0 & 79.3 & 20.9 & 18.1 & 19.4 & 37.3 \\
VideoChat3~\cite{videochat3} & 4B & 74.9 & 23.7 & 0.0 & 14.7 & 33.5 & 17.7 & 0.0 & 20.4 & 49.4 & 26.3 & 22.2 & 24.1 & 31.6 \\
\midrule
\textbf{Qwen3.5-4B} & 4B & 32.0 & 24.6 & 14.3 & 12.4 & 21.0 & 4.8 & 0.0 & 23.0 & 28.2 & 19.2 & 12.6 & 15.6 & 19.4 \\
\quad + JoyAI system & 4B & 30.8 & 15.7 & 35.7 & 26.3 & 38.1 & 19.8 & 0.0 & 10.6 & 23.3 & 34.7 & 21.4 & 27.4 & 26.2 \\
\quad + IPI-Agent & 4B & 75.8 & 64.4 & 0.0 & 1.9 & 10.4 & 4.6 & 0.0 & 1.2 & 70.3 & 6.7 & 5.3 & 5.9 & 25.1 \\
\rowcolor{ocrow}
\quad + \livecue & 4B & 67.1 & 33.6 & 28.6 & 30.1 & 41.5 & 20.8 & 100.0 & 49.2 & 50.5 & 38.0 & 33.1 & 35.3 & 39.8 \\
\midrule
\textbf{Qwen3.6-35B-A3B} & 35B-A3B & 38.2 & 24.5 & 14.3 & 18.8 & 31.1 & 11.1 & 0.0 & 21.5 & 31.5 & 26.8 & 18.5 & 22.3 & 25.0 \\
\quad + JoyAI system & 35B-A3B & 29.6 & 13.1 & 21.4 & 22.2 & 30.4 & 5.4 & 0.0 & 23.5 & 21.3 & 28.2 & 15.9 & 21.5 & 21.5 \\
\quad + IPI-Agent & 35B-A3B & 84.0 & 69.9 & 0.0 & 2.4 & 13.8 & 3.7 & 0.0 & 6.4 & 77.1 & 9.2 & 6.9 & 8.0 & 28.6 \\
\rowcolor{ocrow}
\quad + \livecue & 35B-A3B & 73.0 & 35.0 & 21.4 & 34.0 & 46.7 & 29.6 & 0.0 & 53.7 & 54.2 & 42.9 & 39.1 & 40.9 & 44.8 \\
\midrule
\textbf{Qwen3.8-27B} & 27B & 44.0 & 44.1 & 26.2 & 43.1 & 49.5 & 16.2 & 0.0 & 49.2 & 43.9 & 49.0 & 31.8 & 39.6 & 40.9 \\
\quad + JoyAI system & 27B & 35.8 & 73.4 & 28.6 & 27.0 & 39.4 & 5.3 & 100.0 & 41.8 & 54.3 & 35.9 & 23.1 & 28.9 & 36.5 \\
\quad + IPI-Agent & 27B & 89.3 & 76.0 & 0.0 & 3.1 & 25.9 & 10.5 & 0.0 & 14.3 & 82.6 & 16.4 & 15.9 & 16.1 & 35.9 \\
\rowcolor{ocrow}
\quad + \livecue & 27B & 77.3 & 44.0 & 40.5 & 46.3 & 57.3 & 32.8 & 0.0 & 59.0 & 60.8 & 55.9 & 43.6 & 49.2 & 52.6 \\
\bottomrule
\end{tabular}
\end{adjustbox}
\caption{\textbf{Detailed evaluation results on InteractionBench.} The first row of each VLM group is the VLM used directly (Direct). Capability and A/B/C entries use the official Total score. A, B, and C denote immediate answers, delayed triggers, and repeated reports; B/C combines the latter two; Overall uses the official benchmark aggregate.}
\label{tab:app:ib}
\end{table}

\begin{table}[H]
\centering
\small
\setlength{\tabcolsep}{2pt}
\renewcommand{\arraystretch}{1.12}
\begin{adjustbox}{max width=\linewidth}
\begin{tabular}{lc|ccccc|c}
\toprule
\textbf{Method} & \textbf{Size} & \textbf{SemCor} $\uparrow$ & \textbf{SumFluen} $\uparrow$ & \textbf{TimCover} $\uparrow$ & \textbf{TimDiff} $\downarrow$ & \textbf{TimRedun} $\downarrow$ & \textbf{RNG} $\uparrow$ \\
\midrule
\multicolumn{8}{l}{\textit{Streaming-trained models}} \\
Mage-VL~\cite{mage} & 4B & 0.201 & 0.665 & 35.445 & 2.383 & 0.669 & 15.3 \\
JoyAI-VL-Interaction~\cite{joyai} & 8B & 4.607 & 4.351 & 99.286 & 0.274 & 2.161 & 54.8 \\
AURA~\cite{aura} & 8B & 0.000 & 0.000 & 0.000 & 3.393 & 1.000 & 0.0 \\
MOSS-VL-Realtime~\cite{moss} & 11B & 1.153 & 2.065 & 27.369 & 2.500 & 0.983 & 24.8 \\
VideoChat3~\cite{videochat3} & 4B & 1.905 & 2.973 & 49.804 & 1.712 & 1.127 & 34.5 \\
\midrule
\textbf{Qwen3.5-4B} & 4B & 1.604 & 2.264 & 45.309 & 1.612 & 1.896 & 29.3 \\
\quad + JoyAI system & 4B & 0.788 & 2.317 & 16.771 & 2.937 & 0.878 & 21.1 \\
\quad + IPI-Agent & 4B & 0.714 & 3.022 & 52.606 & 2.142 & 1.117 & 28.0 \\
\rowcolor{ocrow}
\quad + \livecue & 4B & 4.299 & 5.388 & 88.885 & 0.768 & 1.110 & 56.0 \\
\midrule
\textbf{Qwen3.6-35B-A3B} & 35B-A3B & 2.466 & 3.517 & 52.041 & 1.465 & 2.284 & 35.4 \\
\quad + JoyAI system & 35B-A3B & 4.255 & 4.437 & 85.372 & 0.591 & 2.049 & 50.6 \\
\quad + IPI-Agent & 35B-A3B & 2.146 & 5.924 & 37.128 & 2.828 & 0.771 & 37.0 \\
\rowcolor{ocrow}
\quad + \livecue & 35B-A3B & 3.692 & 5.355 & 75.960 & 0.407 & 1.524 & 53.1 \\
\midrule
\textbf{Qwen3.8-27B} & 27B & 1.154 & 1.664 & 22.394 & 2.696 & 1.363 & 21.8 \\
\quad + JoyAI system & 27B & 4.531 & 4.551 & 93.171 & 0.116 & 3.148 & 52.9 \\
\quad + IPI-Agent & 27B & 1.585 & 5.782 & 24.876 & 3.355 & 0.769 & 31.2 \\
\rowcolor{ocrow}
\quad + \livecue & 27B & 4.978 & 5.452 & 95.499 & 0.628 & 1.206 & 59.1 \\
\bottomrule
\end{tabular}
\end{adjustbox}
\caption{\textbf{Detailed evaluation results on OmniStar-RNG.} The first row of each VLM group is the VLM used directly (Direct). SemCor and SumFluen are rated from 0 to 10; TimCover is a percentage, TimDiff is in seconds, and TimRedun is a redundancy measure. RNG uses Eq.~\ref{eq:rng-balance}.}
\label{tab:app:rng}
\end{table}

\FloatBarrier

\section{Further Analysis of General VLMs}
\label{sec:app:analyses}

This section extends the diagnosis of Sec.~\ref{sec:obs:diagnosis}: whether the answer content of general VLMs holds beyond selective answering, how the explicit event confirmation is set up and where it stops, and what streaming adaptation costs trained models offline.

\subsection{Answer Content on Jointly Answered Windows}
\label{sec:app:content}

Table~\ref{tab:content-quality} extends the content comparison of Table~\ref{tab:direct-content} (Sec.~\ref{sec:obs:diagnosis}) to the ground-truth windows answered by Direct, \livecue, and each streaming-trained model, scored by each benchmark's native content judge, as discussed in Sec.~\ref{sec:exp:main}.
\begin{table}[H]
\centering
\footnotesize
\setlength{\tabcolsep}{2.8pt}
\begin{tabular}{@{}lrcrcrcrc@{}}
\toprule
\multirow{2}{*}{Trained model} & \multicolumn{2}{c}{ProactiveVQA (0--2)} & \multicolumn{2}{c}{ViSpeak (0--5)} & \multicolumn{2}{c}{IB non-MCQ (0--1)} & \multicolumn{2}{c}{OmniStar-RNG (0--10)} \\
\cmidrule(lr){2-3}\cmidrule(lr){4-5}\cmidrule(lr){6-7}\cmidrule(l){8-9}
 & $N$ & Direct / \livecue / Trained & $N$ & Direct / \livecue / Trained & $N$ & Direct / \livecue / Trained & $N$ & Direct / \livecue / Trained \\
\midrule
Mage-VL & 122 & 1.57 / 1.37 / 0.84 & 296 & 4.08 / 4.08 / 1.48 & 510 & 0.65 / 0.68 / 0.51 & 101 & 4.49 / 4.60 / 0.40 \\
JoyAI-VL-Interaction & 582 & 1.42 / 1.32 / 1.13 & 343 & 4.05 / 4.10 / 3.78 & 497 & 0.63 / 0.68 / 0.32 & 243 & 5.06 / 5.09 / 4.52 \\
AURA & 634 & 1.43 / 1.38 / 1.20 & 141 & 4.62 / 4.65 / 3.45 & 297 & 0.54 / 0.61 / 0.34 & 0 & -- \\
MOSS-VL-Realtime & 871 & 1.44 / 1.39 / 1.21 & 461 & 4.16 / 4.19 / 3.30 & 405 & 0.69 / 0.73 / 0.33 & 99 & 4.79 / 5.05 / 4.23 \\
VideoChat3 & 693 & 1.48 / 1.37 / 0.94 & 225 & 4.00 / 4.12 / 2.86 & 357 & 0.73 / 0.76 / 0.50 & 148 & 4.79 / 4.87 / 3.77 \\
\bottomrule
\end{tabular}
\caption{\textbf{Answer content of Direct, \livecue, and streaming-trained models on windows answered by all three.} Scores use each benchmark's content judge on the same ground-truth windows; unanswered windows are not scored. \livecue keeps content above every trained model while covering far more windows than Direct; on ProactiveVideoQA its content is slightly below Direct's.}
\label{tab:content-quality}
\end{table}

\begin{figure}[H]
\centering
\includegraphics[width=\textwidth]{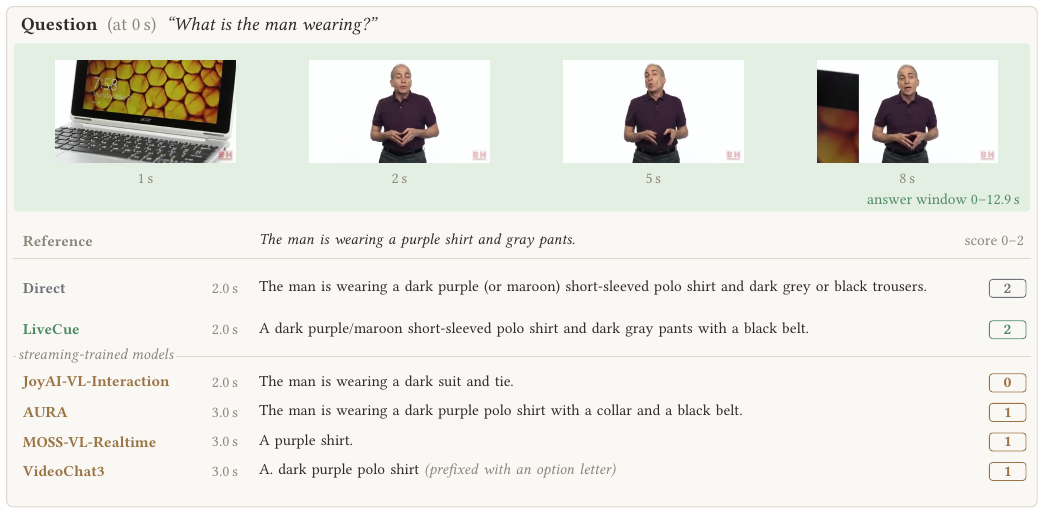}
\caption{\textbf{An illustrative ProactiveVideoQA window} answered by Direct, \livecue (both Qwen3.8-27B), and four streaming-trained models. Each row gives the system's first response inside the answer window and its content score (0--2) from the benchmark's judge. Mage-VL does not respond inside this window.}
\label{fig:app:content-case}
\end{figure}

\paragraph{Content beyond selective answering.}
Because Direct answers only part of the windows, its advantage on jointly answered windows could in principle reflect answering only when the evidence is easy. Two observations make this explanation insufficient. First, the advantage does not shrink as coverage grows: \livecue answers 63.8\% of ProactiveVideoQA windows and 95.1\% of OmniStar-RNG windows, and on the much larger sets of windows it answers together with each trained model, its content remains higher on all four benchmarks (Table~\ref{tab:content-pair}). Second, the trained models with the highest coverage are not the strongest in content: JoyAI-VL-Interaction answers 99.3\% of OmniStar-RNG windows and 68.0\% of the non-multiple-choice InteractionBench windows, yet scores 4.55 and 0.24 on the windows it shares with \livecue, against 5.11 and 0.56. The gap therefore reflects the quality of what is said rather than which windows are answered. Figure~\ref{fig:app:content-case} shows a window answered by Direct, \livecue, and four trained models.
\begin{table}[H]
\centering
\footnotesize
\setlength{\tabcolsep}{3.2pt}
\begin{tabular}{@{}l rrc rrc rrc rrc@{}}
\toprule
\multirow{2}{*}{Trained model} & \multicolumn{3}{c}{ProactiveVQA (0--2)} & \multicolumn{3}{c}{ViSpeak (0--5)} & \multicolumn{3}{c}{IB non-MCQ (0--1)} & \multicolumn{3}{c}{OmniStar-RNG (0--10)} \\
\cmidrule(lr){2-4}\cmidrule(lr){5-7}\cmidrule(lr){8-10}\cmidrule(l){11-13}
 & Cov. & $N$ & \livecue / Trained & Cov. & $N$ & \livecue / Trained & Cov. & $N$ & \livecue / Trained & Cov. & $N$ & \livecue / Trained \\
\midrule
Mage-VL & 7.3 & 162 & 1.36 / 0.75 & 56.2 & 336 & 3.97 / 1.40 & 56.7 & 888 & 0.58 / 0.44 & 38.6 & 371 & 4.45 / 0.53 \\
JoyAI-VL-Interaction & 39.6 & 891 & 1.27 / 1.05 & 56.6 & 399 & 3.88 / 3.58 & 68.0 & 904 & 0.56 / 0.24 & 99.3 & 942 & 5.11 / 4.55 \\
AURA & 55.1 & 885 & 1.33 / 1.16 & 22.2 & 147 & 4.56 / 3.33 & 27.7 & 480 & 0.52 / 0.27 & 0.0 & 0 & -- \\
MOSS-VL-Realtime & 66.0 & 1,646 & 1.33 / 1.10 & 74.1 & 534 & 3.96 / 3.10 & 41.7 & 681 & 0.64 / 0.28 & 26.9 & 265 & 4.81 / 4.16 \\
VideoChat3 & 31.9 & 956 & 1.32 / 0.87 & 32.0 & 248 & 3.92 / 2.72 & 24.7 & 468 & 0.69 / 0.43 & 56.5 & 538 & 4.65 / 3.80 \\
\midrule
\livecue coverage & 63.8 & & & 75.4 & & & 49.0 & & & 95.1 & & \\
Direct coverage & 36.2 & & & 67.8 & & & 29.4 & & & 25.2 & & \\
\bottomrule
\end{tabular}
\caption{\textbf{Answer content of \livecue and streaming-trained models on windows answered by both.} Cov.\ is the percentage of ground-truth windows that each trained model answers, over its valid windows; $N$ is the number of windows answered by both \livecue and the trained model, scored by each benchmark's content judge. The last two rows give the coverage of \livecue and Direct.}
\label{tab:content-pair}
\end{table}

\FloatBarrier

\subsection{Explicit Event Confirmation}
\label{sec:app:confirm}

Table~\ref{tab:app:confirm} details the prefix query in Sec.~\ref{sec:obs:diagnosis}. For each OmniMMI-PA target, a positive prefix ends inside the target window and a paired negative prefix of the same video ends before the event. The VLM receives frames sampled uniformly from the prefix and answers whether the requested event has occurred; all outputs are valid yes/no answers. All three VLMs confirm most target events, and their accuracy exceeds a rule that predicts from prefix length alone. On the 64 pairs that this rule misclassifies, Qwen3.8-27B still answers 49 correctly. Over all 143 visual targets, Qwen3.8-27B confirms 141, including 54 of the 55 targets for which Direct delivers no response inside the window.

\begin{table}[H]
\centering
\footnotesize
\begin{tabular}{@{}lccc@{}}
\toprule
Model & Pos.\ recall & Neg.\ spec. & Acc. \\
\midrule
Duration-only rule & -- & -- & 75.0 \\
Qwen3.5-4B & 87.5 & 89.8 & 88.7 \\
Qwen3.6-35B-A3B & 93.8 & 82.8 & 88.3 \\
Qwen3.8-27B & 98.4 & 77.3 & 87.9 \\
\bottomrule
\end{tabular}
\caption{\textbf{Explicit event confirmation} on 128 paired OmniMMI-PA prefixes (\%). A positive prefix ends inside the target window and its paired negative ends before the event; the model answers whether the requested event has occurred. The duration-only rule is a five-fold, video-level cross-validated threshold on prefix length.}
\label{tab:app:confirm}
\end{table}

\begin{figure}[H]
\centering
\includegraphics[width=\textwidth]{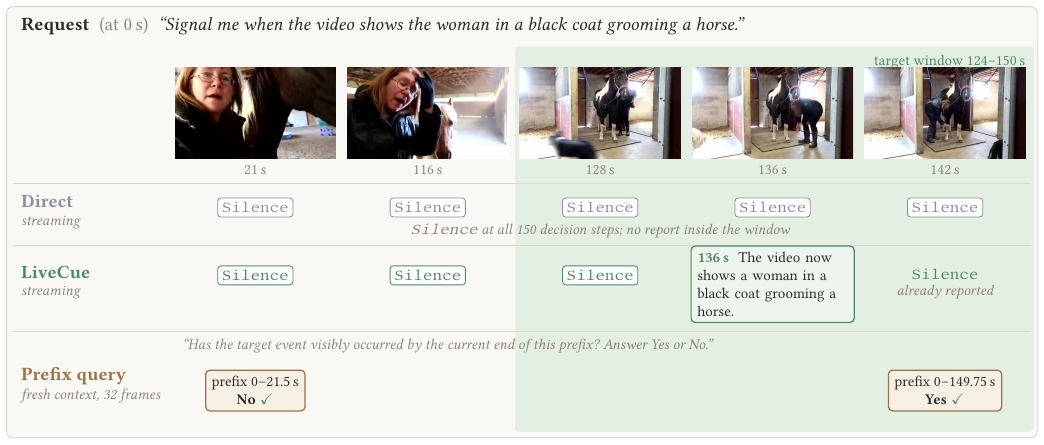}
\caption{\textbf{An illustrative OmniMMI-PA target} with Qwen3.8-27B. Top: frames at the marked video times; the shaded band is the target window. Direct outputs \silence at every decision step. The prefix query, with a fresh context and 32 frames sampled from the prefix, answers No for the prefix ending before the event and Yes for the prefix ending inside the window. \livecue, built on the same frozen model with the same visual input, reports the event once at 136\,s and stays silent afterwards.}
\label{fig:app:pa-case}
\end{figure}

\paragraph{Scope of the query.}
The query differs from streaming inference in three respects: it starts from a fresh context without interaction history, it samples frames from the whole prefix rather than the recent window seen during streaming, and it asks a direct yes/no question about the named event rather than leaving the decision to report to the model. Its positive answers are also read together with the paired negatives, since Qwen3.8-27B rejects 77.3\% of the negative prefixes. The query thus shows that the requested events are recognizable by the same model from the frames observed so far; it does not show that the model had recognized them at a silent streaming tick. A complementary view comes from the Direct-to-\livecue comparison (Table~\ref{tab:abl-ladder}): with the same frozen model and the same visual input, stating the interaction as a protocol and deciding against it raises OmniMMI-PA from 49.5 to 59.0, so a large part of the gap closes without any change to perception. Figure~\ref{fig:app:pa-case} illustrates one such target: Direct stays silent throughout the stream, the prefix query confirms the event inside the window and rejects the paired prefix before it, and \livecue, with the same frozen model, reports the event once inside the window.

\begin{figure}[H]
\centering
\includegraphics[width=\textwidth]{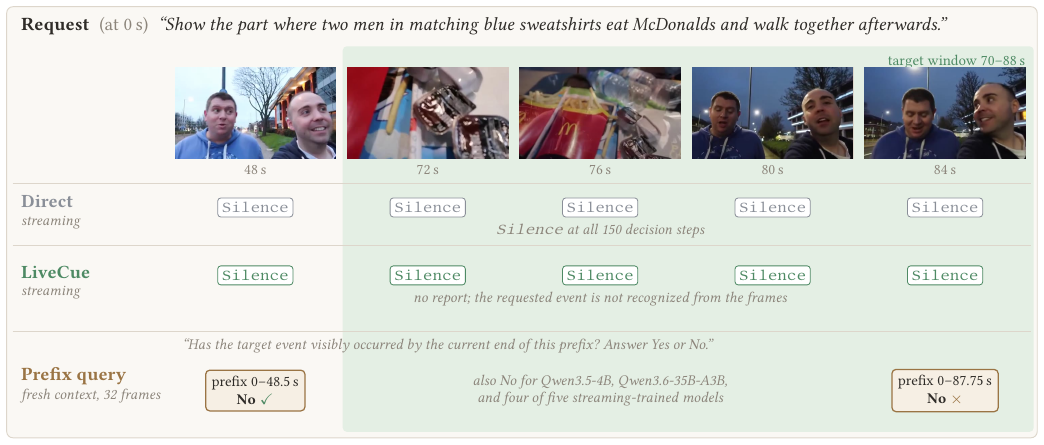}
\caption{\textbf{A target that the prefix query also misses}, with Qwen3.8-27B. The meal is shown only as close-ups of the food, and the walk afterwards takes place in low light, so the combined description in the request is not confirmed from the frames. Direct and \livecue output \silence throughout, and the prefix query answers No both before and inside the window.}
\label{fig:app:pa-fail}
\end{figure}

\noindent The query also delimits what organization can recover. For 2 of the 143 targets, Qwen3.8-27B answers No even on the prefix that ends inside the window; Figure~\ref{fig:app:pa-fail} shows one of them. Direct and \livecue both stay silent, and the same query is also answered No by the other two VLMs and by four of the five streaming-trained models. When the requested event is not recognized from the frames, a different interaction organization cannot supply the missing perception.

\FloatBarrier

\subsection{Offline Understanding after Streaming Adaptation}
\label{sec:app:offline}

Table~\ref{tab:app:offline} collects the offline video-understanding results that streaming methods report for their models and the corresponding starting checkpoints, as noted in Sec.~\ref{sec:related}. \livecue leaves the VLM weights unchanged and therefore retains the offline abilities of the underlying model.
\begin{table}[H]
\centering
\small
\begin{tabular}{llcccc}
\toprule
Backbone / streaming model & Source & Video-MME & LongVideoBench & MVBench & EgoSchema \\
\midrule
Qwen3-VL-8B-Instruct & \multirow{3}{*}{\cite{aura}} & 68.6 & 61.9 & 69.0 & -- \\
AURA & & 65.1 & 58.8 & 68.1 & -- \\
$\Delta$ & & \textcolor{deltadown}{$-$3.5} & \textcolor{deltadown}{$-$3.1} & \textcolor{deltadown}{$-$0.9} & -- \\
\midrule
MOSS-VL-Instruct & \multirow{3}{*}{\cite{joyai}} & -- & 66.12 & 65.62 & 62.80 \\
MOSS-VL-Realtime & & -- & 60.51 & 61.18 & 57.40 \\
$\Delta$ & & -- & \textcolor{deltadown}{$-$5.61} & \textcolor{deltadown}{$-$4.44} & \textcolor{deltadown}{$-$5.40} \\
\midrule
Qwen3-VL & \multirow{3}{*}{\cite{proactvl}} & -- & -- & 66.30 & -- \\
Proact-VL & & -- & -- & 64.65 & -- \\
$\Delta$ & & -- & -- & \textcolor{deltadown}{$-$1.65} & -- \\
\midrule
InternVL-3.5-8B & \multirow{3}{*}{\cite{evostreaming}} & 64.7 & 61.0 & -- & 60.0 \\
EvoStreaming (InternVL-3.5) & & 63.5 & 59.2 & -- & 58.0 \\
$\Delta$ & & \textcolor{deltadown}{$-$1.2} & \textcolor{deltadown}{$-$1.8} & -- & \textcolor{deltadown}{$-$2.0} \\
\midrule
Qwen2.5-VL-7B & \multirow{3}{*}{\cite{evostreaming}} & 62.0 & 58.3 & -- & 63.2 \\
EvoStreaming (Qwen2.5-VL) & & 61.9 & 58.1 & -- & 58.9 \\
$\Delta$ & & \textcolor{deltadown}{$-$0.1} & \textcolor{deltadown}{$-$0.2} & -- & \textcolor{deltadown}{$-$4.3} \\
\midrule
Qwen2-VL-7B & \multirow{3}{*}{\cite{evostreaming}} & 57.9 & 55.5 & -- & 62.2 \\
EvoStreaming (Qwen2-VL) & & 57.9 & 53.4 & -- & 57.9 \\
$\Delta$ & & 0.0 & \textcolor{deltadown}{$-$2.1} & -- & \textcolor{deltadown}{$-$4.3} \\
\midrule
StreamBridge (general video data)$^\dagger$ & \multirow{3}{*}{\cite{streambridge}} & 61.7 & -- & 64.5 & -- \\
StreamBridge (Stream-IT only) & & 59.0 & -- & 58.8 & -- \\
$\Delta$ & & \textcolor{deltadown}{$-$2.7} & -- & \textcolor{deltadown}{$-$5.7} & -- \\
\bottomrule
\end{tabular}
\caption{\textbf{Offline video understanding before and after streaming adaptation, as reported in prior work.} Scores follow each study's own evaluation settings, and comparisons are made only within each group. $\Delta$ is the streaming model's score minus the corresponding baseline, in percentage points; -- marks results not reported. MOSS-VL numbers are taken from the JoyAI-VL-Interaction report. $^\dagger$StreamBridge compares training-data recipes on the same backbone rather than an original checkpoint with its adapted version. Proact-VL additionally includes gaming-domain adaptation. Sources: AURA Tab.~5; JoyAI-VL-Interaction Tab.~4; Proact-VL Tab.~19; EvoStreaming Tab.~3; StreamBridge Tab.~5.}
\label{tab:app:offline}
\end{table}

\FloatBarrier

\section{Protocol Library and Prompts}
\label{sec:app:protocols}

\begin{figure}[H]
\centering
\includegraphics[width=\textwidth]{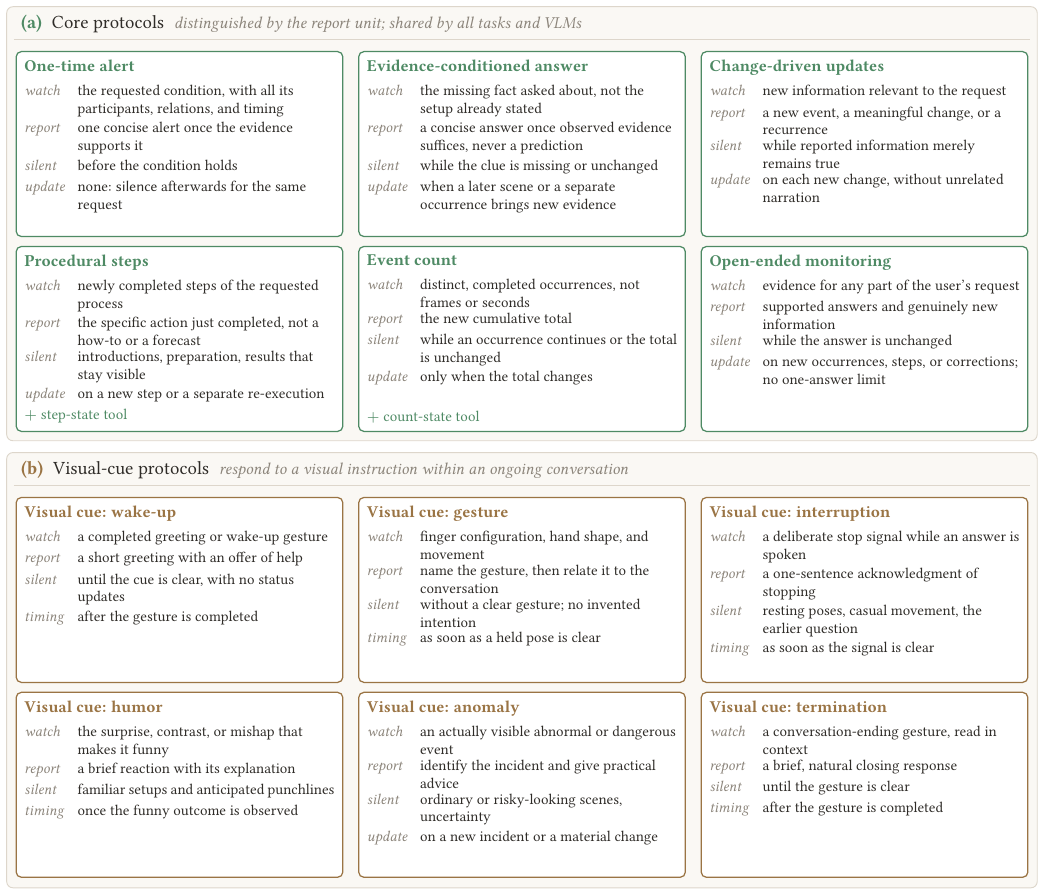}
\caption{\textbf{The protocol library.} Each card summarizes what the protocol directs the model to watch, what forms one report, when to stay silent, and when to update or respond. (a) Core protocols, named by their report unit and shared by all tasks; the procedural-step and event-count protocols are paired with the step and count tools of Sec.~\ref{sec:method:state}. (b) Visual-cue protocols for responding to visual instructions in a conversation.}
\label{fig:app:protocols}
\end{figure}

\paragraph{Sources.}
The protocol library is distilled from the task definitions and illustrative examples given in the papers of existing proactive and streaming benchmarks. For each task, we ask three questions: what evidence the user expects the assistant to watch for, what constitutes one report, and when the assistant should wait or report again. Tasks that share these answers share a protocol, which yields the six core protocols of Figure~\ref{fig:app:protocols}(a), each named by its report unit. A protocol fixes only this interaction form; the objects, events, and conditions always come from the user's request at run time.

\paragraph{Held-out evaluation.}
InteractionBench~\cite{interactionbench}, whose tasks span alerts, repeated reports, delayed triggers, counting, and negative items, was not consulted in this process, neither its task definitions nor its examples, and serves only as an evaluation suite. Its requests are routed to the same core protocols without any InteractionBench-specific protocol, and \livecue still improves over Direct on all three VLMs (by 20.4, 19.8, and 11.7 points) and reaches 52.6 with Qwen3.8-27B, above every streaming-trained model and agent framework (Table~\ref{tab:main-results}).

\paragraph{Visual-cue protocols.}
Some requests ask the assistant not to watch for an event but to respond to a visual instruction within an ongoing conversation, such as a wake-up gesture, a stop signal, or a humorous moment. Each of these response forms is described by a visual-cue protocol (Figure~\ref{fig:app:protocols}(b)). They are selected by the same router and run with the same execution loop, output interface, and commitment as the core protocols; only the protocol card differs.

\paragraph{Prompt structure.}
Every decision call uses one system prompt built from three parts shared across all VLMs. A fixed preamble tells the model that it is following a timestamped video stream for the user's active request, that at each update it should decide from the observations and its own response history whether a response is warranted, and that it should otherwise output \silence without predicting unobserved events. The selected protocol card follows. A closing rule sets the response timing: most protocols wait until a requested action is observed to be complete or an answer is supported, while the gesture and interruption cues allow a response as soon as the cue is clearly occurring. The request and the timestamped frames are given as user messages.

\paragraph{Protocol selection.}
At task activation, the same VLM receives the request and the visible dialogue context in one text-only call, without video, together with a one-line description of each protocol, and returns a single protocol identifier as JSON (Eq.~\ref{eq:configure}). It is asked to identify the active instruction and to choose by the intended behavior, preserving scope, negation, participants, temporal conditions, and reporting frequency, rather than by keywords. When the request leaves the reporting frequency unspecified, application defaults apply: relevant updates for changing activities or states, cumulative updates for totals, and a single notification for a target condition. Explicit scope and frequency in the request override these defaults, and open-ended monitoring is chosen when no narrower protocol faithfully covers the request.

\FloatBarrier

\section{Ablations and Robustness}
\label{sec:app:ablation}

We first check that the largest step of the Direct-to-\livecue ladder is not an artifact of output parsing, then compare routing with fixed protocols and test its stability to rewording, and finally report the statistics behind the task-state and pacing ablations.

\subsection{Output Parsing of Direct}
\label{sec:app:parser}

The largest step of Table~\ref{tab:abl-ladder}, from Direct with the \silence/\response interface (38.8) to the general protocol with explicit decisions (50.5), also changes the output format, so part of it could come from parsing rather than from the protocol. We therefore keep the Direct outputs, histories, and stopping points fixed and re-parse them under cumulative recovery rules fixed in advance: surface variants of the tags, answers written outside the tag, answers without a tag, and tag headers missing their closing bracket (Table~\ref{tab:app:parser}). A further rule that reads explicit statements of silence changes no output. Every recovered response is kept, whether or not its timing is correct. With all rules, invalid outputs of the \silence/\response variant fall from 11.8\% to none and 15,275 responses are recovered, raising Overall from 38.8 to 40.6, mainly on OVO-Timing (19.4 to 24.0) and OmniStar-RNG (21.8 to 27.1). The recovered responses are not better timed than the rest: OVO-Timing recall rises from 21.4\% to 33.6\% while precision stays near 20\%, and OmniMMI-PA first-trigger accuracy falls from 49.5 to 49.0 because some recovered answers precede the event. The free-form variant is unchanged (27.1 to 27.0). Even under the most tolerant parsing, Direct remains 9.9 points below the general protocol with explicit decisions, so parsing accounts for at most 1.8 of the 11.7 points; the rest comes from stating the interaction as a protocol and deciding against it at every tick.

\begin{table}[H]
\centering
\footnotesize
\setlength{\tabcolsep}{4pt}
\begin{tabular}{@{}lccccccc@{}}
\toprule
System and parsing & PVQA & OVO & PA & ViSp. & IB & RNG & Overall \\
\midrule
\multicolumn{8}{@{}l}{\textcolor{mute}{\textit{Direct, free-form}}} \\
Strict & 51.3 & 8.1 & 0.0 & 14.6 & 34.0 & 54.6 & 27.1 \\
All recovery rules & 50.8 & 8.1 & 0.0 & 14.6 & 33.9 & 54.7 & 27.0 \\
\midrule
\multicolumn{8}{@{}l}{\textcolor{mute}{\textit{Direct, \silence/\response}}} \\
Strict & 41.9 & 19.4 & 49.5 & 59.1 & 40.9 & 21.8 & 38.8 \\
+ Tag surface variants & 41.9 & 19.4 & 49.5 & 59.1 & 40.9 & 21.8 & 38.8 \\
+ Answer outside the tag & 42.8 & 23.6 & 49.0 & 59.1 & 41.0 & 26.3 & 40.3 \\
+ Untagged answer & 43.0 & 24.0 & 49.0 & 59.2 & 41.1 & 27.1 & 40.5 \\
+ Incomplete tag header & 43.0 & 24.0 & 49.0 & 59.3 & 41.1 & 27.1 & 40.6 \\
\midrule
\rowcolor{ocrow}+ Protocol, decision & \textbf{47.3} & \textbf{32.6} & \textbf{59.0} & \textbf{59.9} & \textbf{49.3} & \textbf{54.7} & \textbf{50.5} \\
\bottomrule
\end{tabular}
\caption{\textbf{Tolerant parsing of Direct} on Qwen3.8-27B. The model outputs are fixed and only re-parsed; recovery rules are added cumulatively, and every recovered response is kept regardless of its timing. The last row is the general protocol with explicit decisions from Table~\ref{tab:abl-ladder}.}
\label{tab:app:parser}
\end{table}

\FloatBarrier

\subsection{Routing versus Fixed Protocols}
\label{sec:app:protocol}

We compare routing with each of the six core protocols fixed for all tasks; ViSpeak, whose requests concern visual cues within a conversation, is excluded (Table~\ref{tab:abl-protocol}). A fixed protocol keeps the state support it has in \livecue, so the procedural-step and event-count protocols run with their tools, and the routed system is the full \livecue. The comparison covers procedural steps, clue-triggered answers, counting, repeated reports, negative items, and continuous narration. Each fixed protocol remains strong where its report unit matches the task. With its tool, the procedural-step protocol leads on EGO, CRR, and negative items and matches routing on SSR, and the event-count protocol raises REC to 52.0 and the InteractionBench counting score from 20.0 to 28.5. None of them, however, serves every task. Routing obtains the best mean rank (1.5 against 2.1 for the next system) and the best Macro-5 (52.4 against 50.9). It leads on REC, repeated reports, and OmniStar-RNG, and at the benchmark level on OVO-Timing and OmniMMI-PA (Table~\ref{tab:app:protocol}).

The event-count protocol shows why state support follows the protocol rather than being given to every task. Its counting record improves counting, yet imposed on all tasks it leaves the narration of OmniStar-RNG almost silent: the model says nothing on 185 of the 200 videos and reports only numbers on the rest, so the RNG score falls from 56.8 without the tool to 0.0. This agrees with the state ablation in the main text, where supplying state to every task lowers the tasks with independent reports.
\begin{table}[H]
\centering
\footnotesize
\setlength{\tabcolsep}{1.3pt}
\begin{tabular}{@{}lccccccccc@{}}
\toprule
 & PVQA & \multicolumn{3}{c}{OVO-Timing} & \multicolumn{2}{c}{IB} & & & \\
\cmidrule(lr){2-2}\cmidrule(lr){3-5}\cmidrule(lr){6-7}
Protocol & EGO & CRR & SSR & REC & Rep. & Neg. & RNG & Rank & Macro-5 \\
\midrule
\multicolumn{10}{@{}l}{\textcolor{mute}{\textit{Fixed for all tasks}}} \\
Monitor & 37.4 & 31.7 & 24.5 & 39.5 & 37.1 & 58.0 & \textbf{59.1} & 4.8 & 49.5 \\
Alert & 37.7 & 31.4 & 15.7 & 31.4 & 34.5 & 68.1 & 53.4 & 6.1 & 47.0 \\
Evidence answer & 34.0 & 35.3 & 17.9 & 40.6 & 36.7 & 73.9 & 57.0 & 4.7 & 47.3 \\
Change-driven & 37.9 & 33.8 & 25.7 & 39.5 & 38.2 & 69.6 & 58.9 & 4.0 & 50.0 \\
Steps + step tool & \textbf{45.4} & \textbf{38.8} & \textbf{32.9} & 47.3 & 43.4 & \textbf{85.5} & 55.9 & 2.1 & 50.9 \\
Count + count tool & 35.2 & 37.8 & 17.3 & 52.0 & 41.6 & 66.7 & 0.0$^\dagger$ & 4.7 & 38.1 \\
\midrule
\rowcolor{ocrow}Routed (\livecue) & 43.6 & 37.9 & \textbf{32.9} & \textbf{52.5} & \textbf{43.6} & 76.8 & \textbf{59.1} & \textbf{1.5} & \textbf{52.4} \\
\bottomrule
\end{tabular}
\caption{\textbf{Routing versus fixed protocols} on Qwen3.8-27B. Each protocol is fixed for all tasks together with the state support it has in \livecue: the procedural-step and event-count protocols use their tools, and the other protocols use none. EGO: ProactiveVideoQA-EGO pAUC; CRR, SSR, REC: OVO-Timing F1; Rep.\ and Neg.: InteractionBench repeated-report items and negative items; Rank: mean rank among the seven systems over the seven task columns; Macro-5: average of the five benchmarks other than ViSpeak. $^\dagger$Silent on 185 of the 200 OmniStar-RNG videos and reports only numbers on the rest.}
\label{tab:abl-protocol}
\end{table}

\begin{table}[H]
\centering
\footnotesize
\setlength{\tabcolsep}{3pt}
\begin{tabular}{@{}lcccccc@{}}
\toprule
Protocol & PVQA & OVO & PA & IB & RNG & Macro-5 \\
\midrule
Open-ended monitoring & 47.6 & 31.9 & 59.0 & 49.9 & \textbf{59.1} & 49.5 \\
One-time alert & 44.0 & 26.2 & 61.0 & 50.5 & 53.4 & 47.0 \\
Evidence-conditioned answer & 43.5 & 31.3 & 57.0 & 47.7 & 57.0 & 47.3 \\
Change-driven updates & 46.5 & 33.0 & 60.0 & 51.9 & 58.9 & 50.0 \\
Procedural steps + step tool & 47.3 & 39.7 & 58.0 & \textbf{53.7} & 55.9 & 50.9 \\
Event count + count tool & 43.1 & 35.7 & 59.0 & 52.8 & 0.0 & 38.1 \\
\midrule
\rowcolor{ocrow}Model routing (\livecue) & \textbf{47.9} & \textbf{41.1} & \textbf{61.5} & 52.6 & \textbf{59.1} & \textbf{52.4} \\
\bottomrule
\end{tabular}
\caption{\textbf{Benchmark-level results of the routing comparison} (Table~\ref{tab:abl-protocol}) on Qwen3.8-27B. PA reports first-trigger accuracy.}
\label{tab:app:protocol}
\end{table}

\subsection{Stability of Protocol Selection}
\label{sec:app:stability}

Protocol selection is a single text-only call, so we test whether it follows the meaning of a request rather than its wording. Each of the 3,036 distinct requests is rewritten in four ways that keep its meaning: a polite wrapper, an unrelated opening sentence, a casual form without capitalization or punctuation, and a paraphrase. The paraphrase is written by Qwen-Max, a model outside the evaluated VLMs, which is instructed to keep the target, conditions, negation, time scope, and reporting frequency. Only the request sentence is rewritten; answer options, answer-format instructions, and the conversation context of ViSpeak stay unchanged. Every rewritten request is routed by Qwen3.8-27B with the prompt and decoding of the main experiments, which reproduce all 3,036 original selections. Table~\ref{tab:app:routing-stability} reports the share of reassigned cases and Macro-6 recomputed with the rewritten routes. A case whose protocol is unchanged keeps its trajectory, and a case moved to another protocol without task state uses that protocol's trajectory from the fixed-protocol runs above. The few remaining cases are bounded by the lowest and highest attainable scores. Rewording reassigns 3.3\% to 8.0\% of the cases, and none on OmniStar-RNG or ViSpeak. The largest share, 10.8\% to 25.3\%, falls on the retrospective questions of InteractionBench A, which can be answered either once supported or through later updates. Macro-6 stays between 53.7 and 54.4 under every rewrite, against 54.2 with the original requests.

\begin{table}[H]
\centering
\small
\setlength{\tabcolsep}{6pt}
\begin{tabular}{@{}lp{0.52\linewidth}cc@{}}
\toprule
Rewrite & Example or rule & Reassigned (\%) & Macro-6 \\
\midrule
Original & -- & -- & 54.2 \\
Polite wrapper & ``Could you please help me with this? \ldots\ Thank you so much!'' & 5.2 & 54.0--54.2 \\
Unrelated opening & ``I just got back from lunch, so let's get started. \ldots'' & 7.7 & 54.0--54.4 \\
Casual form & Lowercase, punctuation removed & 3.3 & 54.1--54.2 \\
Paraphrase & Rewritten by Qwen-Max with the meaning kept & 8.0 & 53.7--54.2 \\
\bottomrule
\end{tabular}
\caption{\textbf{Stability of protocol selection to rewording} (\livecue, Qwen3.8-27B, all 3,859 cases). Reassigned: share of cases routed to another narrower protocol. Macro-6 is recomputed with the rewritten routes; ranges bound the few cases moved to a protocol without a recorded trajectory.}
\label{tab:app:routing-stability}
\end{table}

\FloatBarrier

\subsection{Task State and Pacing}
\label{sec:app:state}

For the task-state ablation (Table~\ref{tab:abl-state}), the tools act only on the tasks routed to procedural steps and event counts, and the step and count records are written from delivered responses alone. Supplying state to every task also lowers OmniMMI-PA event F1 from 69.4 to 62.6.

\paragraph{Response pacing.} With the full \livecue, removing the one-tick cooldown ($k=0$) lets the model answer nearly identical consecutive observations again: delivered responses grow by 60\% (16.6k to 26.7k) and video-inference calls from 284.4k to 300.0k, the InteractionBench redundancy penalty rises from 4.61 to 10.56 and OmniStar-RNG temporal redundancy from 1.21 to 2.90, and RNG semantic correctness falls from 4.98 to 4.84, lowering the RNG score from 59.1 to 54.7. OmniMMI-PA first-trigger accuracy and ViSpeak are unchanged, as pacing acts only after a delivery.

\section{Comparison with IPI-Agent}
\label{sec:app:ipi}

IPI-Agent~\cite{ipibench} is the agent framework closest to \livecue: it also builds proactive interaction around a frozen model at inference time. This section describes how we run it and analyzes where the two designs differ.

\paragraph{Implementation.}
The IPI-Agent code has not been released; the official repository provides the IPIBench data, baselines, and evaluators, and lists the agent as forthcoming. We therefore reimplement IPI-Agent from its paper~\cite{ipibench}, following each component it describes (Table~\ref{tab:app:ipi}). The paper evaluates IPI-Agent with several main models, all paired with the same Qwen3-VL-Embedding-2B gate; we keep this gate and replace the main model with each of our three VLMs, which serves as both router and responder. Details the paper leaves open are fixed once for all benchmarks and VLMs and recorded with every run.

\begin{table}[H]
\centering
\small
\renewcommand{\arraystretch}{1.15}
\begin{tabular}{@{}p{0.27\linewidth}p{0.69\linewidth}@{}}
\toprule
Component in~\cite{ipibench} & Our reimplementation \\
\midrule
Intent Router (Sec.~4.2) & One call of the main VLM per user input, classifying it as a reactive query, a proactive instruction, or a management instruction (add, modify, or cancel a monitoring task), and returning structured actions. A proactive task carries a repeat flag. \\
Proactive and Interaction Memory & Monitoring tasks are stored with their instruction and repeat flag; the interaction memory holds user inputs and delivered responses (most recent 64 entries). Candidates suppressed by the gate are never written to memory. \\
Reactive path & Answered immediately by the Response Tool from the recent frames and the interaction memory; no monitoring task is created. \\
Response Tool (Sec.~4.3) & For each active task at each step, the main VLM produces a speak/silence candidate and its text from the task instruction, the interaction memory, and the most recent 16 frames. \\
Gating Tool (Sec.~4.3, 5.4) & Qwen3-VL-Embedding-2B, as in all IPI-Agent experiments, embeds the frame window and $M=3$ textual proposals of each task generated by the router. With $\Delta_t=\max_m\,(s^m_t-s^m_{t-1})$, a candidate response is suppressed if $\Delta_t<\theta_{\mathrm{low}}$, and a missed response is recovered by an extra Response call if $\Delta_t>\theta_{\mathrm{high}}$. \\
Parallel monitoring & Active tasks are processed in parallel against the same memory snapshot and share one video embedding per step; their outputs are committed together. \\
Stream (Sec.~5.1) & 1\,FPS input and a decision every second; frames up to and including the current time. \\
\midrule
Unspecified in~\cite{ipibench} & Prompts, JSON schema, $M=3$, $\theta_{\mathrm{low}}/\theta_{\mathrm{high}}=0/0.05$, memory budget, temperature 0 without thinking, and the completion rule of one-time tasks: a task completes only when a delivered response confirms the requested event, so a negative or uncertain recovered response does not end monitoring. \\
\bottomrule
\end{tabular}
\caption{\textbf{IPI-Agent reimplementation.} Each component described in the IPI-Agent paper and its realization in our runs. The same implementation is used with all three VLMs.}
\label{tab:app:ipi}
\end{table}

\paragraph{Evaluation.}
IPI-Agent receives the same causal streams, benchmark questions, scorers, and content judge as every other system. Each question is passed verbatim, without any benchmark-derived hint about the request type: whether a request is reactive or proactive, and whether it repeats, is decided by the router itself. A malformed model output ends the affected stream, responses delivered before it are scored, and the rest of the stream counts as silence. Such failures occur on 3 of 3,859 streams with Qwen3.8-27B, 89 with Qwen3.6-35B-A3B, and 422 with Qwen3.5-4B.

\paragraph{Where the difference arises.}
IPI-Agent fixes the interaction type when a request arrives. A request routed as reactive is answered at once and closed, whereas \livecue's evidence-conditioned answer protocol leaves the timing to the model: it answers immediately when the frames already suffice and waits otherwise. The first design suits InteractionBench A, where every question is to be answered at the moment it is asked. With Qwen3.8-27B, the router labels all 316 A questions reactive and answers each exactly once at the question time, which earns the full timing credit and gives IPI-Agent the highest A scores (Table~\ref{tab:app:ipi-reactive}). \livecue scores lower there, yet above JoyAI-system on every VLM and, already with Qwen3.5-4B, above the streaming-trained JoyAI-VL-Interaction, AURA, and VideoChat3. The same decision, however, is applied to requests whose evidence has not yet appeared: the router also labels 1,412 of the 1,427 ProactiveVideoQA requests, 46 of the 48 OVO-Timing CRR questions, 30 of the 42 SSR requests, and over 40\% of the InteractionBench B/C requests reactive, so they are answered before the relevant event happens. Where answering requires waiting for evidence or reporting repeatedly, \livecue leads by a wide margin: on Qwen3.8-27B it reaches 55.9 against 16.4 on delayed triggers, 43.6 against 15.9 on repeated reports, and 37.9 against 0.0 CRR F1.

\begin{table}[H]
\centering
\small
\setlength{\tabcolsep}{5pt}
\begin{tabular}{@{}l|ccc|ccc|ccc|ccc@{}}
\toprule
 & \multicolumn{3}{c|}{IB-A (immediate)} & \multicolumn{3}{c|}{IB-B (delayed)} & \multicolumn{3}{c|}{IB-C (repeated)} & \multicolumn{3}{c}{OVO-CRR F1} \\
System & 4B & 35B & 27B & 4B & 35B & 27B & 4B & 35B & 27B & 4B & 35B & 27B \\
\midrule
JoyAI system & 23.3 & 21.3 & 54.3 & 34.7 & 28.2 & 35.9 & 21.4 & 15.9 & 23.1 & 15.4 & 11.6 & 5.7 \\
IPI-Agent & \textbf{70.3} & \textbf{77.1} & \textbf{82.6} & 6.7 & 9.2 & 16.4 & 5.3 & 6.9 & 15.9 & 2.5 & 1.5 & 0.0 \\
\rowcolor{ocrow}\livecue & 50.5 & 54.2 & 60.8 & \textbf{38.0} & \textbf{42.9} & \textbf{55.9} & \textbf{33.1} & \textbf{39.1} & \textbf{43.6} & \textbf{23.5} & \textbf{29.4} & \textbf{37.9} \\
\bottomrule
\end{tabular}
\caption{\textbf{Immediate versus evidence-dependent answering.} InteractionBench Total on immediate questions (A), delayed triggers (B), and repeated reports (C), and OVO-Timing CRR F1, for the agent frameworks and \livecue on Qwen3.5-4B, Qwen3.6-35B-A3B, and Qwen3.8-27B. For reference, the streaming-trained JoyAI-VL-Interaction, AURA, and VideoChat3 obtain 47.8, 37.0, and 49.4 on A, 30.5, 32.4, and 26.3 on B, 19.0, 20.1, and 22.2 on C, and 14.9, 6.5, and 31.4 CRR F1.}
\label{tab:app:ipi-reactive}
\end{table}

\paragraph{Request type versus gate thresholds.}
Two controls with Qwen3.8-27B separate the request-type decision from our choices for the unspecified settings (Table~\ref{tab:app:ipi-controls}). They use a fixed subset of 100 InteractionBench B and 100 C items, sampled with a fixed seed, together with all 48 CRR and 42 SSR questions of OVO-Timing. On this subset the main setting closely follows its full-set scores, 17.9 and 14.2 against 16.4 and 15.9 on B and C. The first control varies the gate thresholds over $\theta_{\mathrm{low}}\in\{0,0.02\}$ and $\theta_{\mathrm{high}}\in\{0.02,0.05,0.1\}$. The best setting raises B and C by at most 5.8 and 3.8 points while lowering SSR F1, and CRR F1 stays at 0.0 under every setting, because the questions are still routed as reactive and answered before the event. The second control keeps every setting and replaces only the router's request type with the type given by the task definition. Each request is treated as a proactive instruction that repeats where the task asks for repeated reports, and the router still writes the task instruction and its proposals. On this subset the router had labeled 187 of the 290 requests reactive. With the given type, CRR F1 rises from 0.0 to 32.7 and SSR F1 from 10.7 to 23.0, and B and C reach 28.7 and 31.6. The request type is thus the main source of the difference, in line with the routing statistics above. The remaining distance to \livecue (55.9, 43.6, and 37.9 on the full sets) lies in deciding, once a task is monitored, when the observed evidence suffices.

\begin{table}[H]
\centering
\small
\setlength{\tabcolsep}{6pt}
\begin{tabular}{@{}llcccc@{}}
\toprule
Request type & $\theta_{\mathrm{low}}/\theta_{\mathrm{high}}$ & IB-B & IB-C & CRR F1 & SSR F1 \\
\midrule
Decided by the router & 0/0.05 (main runs) & 17.9 & 14.2 & 0.0 & 10.7 \\
 & 0/0.02 & 9.1 & 10.7 & 0.0 & 8.0 \\
 & 0/0.1 & 23.7 & 18.0 & 0.0 & 5.6 \\
 & 0.02/0.02 & 8.6 & 8.7 & 0.0 & 7.5 \\
 & 0.02/0.05 & 16.9 & 12.0 & 0.0 & 8.5 \\
 & 0.02/0.1 & 21.6 & 17.7 & 0.0 & 8.5 \\
\midrule
Given by the task definition & 0/0.05 & \textbf{28.7} & \textbf{31.6} & \textbf{32.7} & \textbf{23.0} \\
\bottomrule
\end{tabular}
\caption{\textbf{Controls for IPI-Agent with Qwen3.8-27B.} Gate thresholds and the source of the request type are varied on a fixed subset: 100 InteractionBench B and 100 C items sampled with a fixed seed, and all CRR and SSR questions of OVO-Timing. All other settings are those of the main runs. When the request type is given, the router still writes the task instruction and its proposals.}
\label{tab:app:ipi-controls}
\end{table}

\FloatBarrier

\section{Efficiency and Latency}
\label{sec:app:efficiency}

\paragraph{Setup.}
We measure inference on NVIDIA H200 GPUs with one active video per system. Three source videos are drawn from each of the six benchmarks, giving 18 videos and 748 decision polls; six of them are additionally replayed in real time, together with two continuous 300-second streams and the 60-second monitoring prefix of an earlier measurement. Every system receives the same causal video stream with its native visual, history, and decoding policies. \livecue and Direct enable CUDA Graph and prefix caching; \livecue receives four frames per second, retains at most 64 frames, and makes decisions at 1\,Hz. System TTFT starts when a decoded frame becomes available and ends at the first decision token of the main model; the complete decision covers input processing, routing and state handling, generation, and delivery. Source-video decoding, network transport, and speech synthesis are outside the measured latency.

\paragraph{Inference efficiency.}
Table~\ref{tab:app:efficiency} compares complete systems. On Qwen3.8-27B, \livecue runs on the same single GPU with the same 66.6\,GB as Direct and adds only about 20 context tokens per call for the protocol and task state. Its mean decision time of 0.58\,s leaves ample margin within the 1\,Hz interval, and its P95 of 1.01\,s is 46\% lower than that of Direct (1.85\,s); the cooldown also removes main-model inference on 49 of the 748 polls. On Qwen3.5-4B, \livecue needs 20.3\,GB and 0.19\,s per decision (P95 0.35\,s), whereas the 11B MOSS-VL-Realtime reaches 148.5\,GB with a context of 26k tokens and a P95 above one second. JoyAI-system needs a second GPU for its memory model and twice the context of \livecue; on each 300-second stream its memory model is invoked once on 764 frames and takes about 47\,s. IPI-Agent calls its main model on only 245 of the 748 polls, so its low mean decision time reflects skipped inference rather than faster calls, whose TTFT is 0.74\,s.
\begin{table}[H]
\centering
\small
\setlength{\tabcolsep}{5pt}
\begin{tabular}{@{}lcrrrcc@{}}
\toprule
System & GPUs & Memory (GB)\metricdown & Context tokens\metricdown & TTFT (s)\metricdown & Decision mean / P95 (s)\metricdown & Model calls \\
\midrule
MOSS-VL-Realtime (11B) & 1 & 148.5 & 26,335 & 0.150 & 0.212 / 1.030 & 748 / 748 \\
Direct (Qwen3.8-27B) & 1 & 66.6 & 5,603 & 0.428 & 0.641 / 1.849 & 748 / 748 \\
JoyAI-system (Qwen3.8-27B) & 2 & 89.6 & 11,694 & 0.293 & 0.632 / 1.182 & 748 / 748 \\
IPI-Agent (Qwen3.8-27B)$^\dagger$ & 2 & 72.0 & 1,128 & 0.741 & 0.266 / 0.541 & 245 / 748 \\
\rowcolor{ocrow}\livecue (Qwen3.5-4B) & 1 & 20.3 & 5,592 & 0.168 & 0.190 / 0.349 & 698 / 748 \\
\rowcolor{ocrow}\livecue (Qwen3.8-27B) & 1 & 66.6 & 5,625 & 0.447 & 0.576 / 1.007 & 699 / 748 \\
\bottomrule
\end{tabular}
\caption{\textbf{Inference efficiency} on 18 videos, three from each benchmark (748 decision polls), with one active video per system on NVIDIA H200 GPUs. Memory is the peak simultaneous usage on the assigned GPUs, including auxiliary models and caches. Context tokens and TTFT are means over main-model calls; TTFT includes the \silence or control prefix. Decision times are video-averaged means and P95 over all polls, including polls without main-model inference; the last column gives polls with a main-model call. $^\dagger$Our reimplementation.}
\label{tab:app:efficiency}
\end{table}

\paragraph{Real-time playback.}
We further release frames on the real video clock and preserve every poll (Table~\ref{tab:app:realtime}). \livecue on Qwen3.5-4B completes all 959 decisions within one second, with a mean below 0.2\,s. On Qwen3.8-27B, all decisions on the 60-second prefix and the six benchmark videos finish within one second, and over the two 300-second streams 581 of 600 do; 18 of the 19 slower polls occur on the InteractionBench stream, which triggers 21 responses, and the queue recovers after each and is empty at the end of every stream.
\begin{table}[H]
\centering
\small
\begin{tabular}{@{}llccc@{}}
\toprule
Stream & System & Latency mean / P95 / max (s)\metricdown & Polls within 1\,s & Responses \\
\midrule
60-second monitoring prefix & \livecue (Qwen3.5-4B) & 0.190 / 0.216 / 0.218 & 60 / 60 & -- \\
 & \livecue (Qwen3.8-27B) & 0.675 / 0.826 / 0.827 & 60 / 60 & -- \\
\midrule
Six benchmark videos & \livecue (Qwen3.5-4B) & 0.190 / 0.230 / 0.816 & 299 / 299 & 5 \\
 & \livecue (Qwen3.8-27B) & 0.650 / 0.840 / 0.963 & 299 / 299 & 10 \\
\midrule
Two 300-second streams & \livecue (Qwen3.5-4B) & 0.178 / 0.284 / 0.759 & 600 / 600 & 150 \\
 & \livecue (Qwen3.8-27B) & 0.793 / 0.831 / 1.795 & 581 / 600 & 22 \\
\bottomrule
\end{tabular}
\caption{\textbf{Real-time playback.} Frames are released on the real video clock and every 1\,Hz decision poll is preserved; latency runs from frame release to the finished decision and includes queuing and response generation. The six benchmark videos are one per benchmark; the 300-second streams are a ProactiveVideoQA-EGO video and an InteractionBench video.}
\label{tab:app:realtime}
\end{table}

\FloatBarrier

\end{document}